\documentclass[10pt,letterpaper,english,amscd,amsmath,amssymb,verbatim,floatfix,nofootinbib,eqsecnum]{article}
\usepackage[T1]{fontenc}
\usepackage[latin9]{inputenc}
\usepackage{babel}

\usepackage{graphicx}
\usepackage{esint}
\usepackage[unicode=true, pdfusetitle,
 bookmarks=true,bookmarksnumbered=false,bookmarksopen=false,
 breaklinks=false,pdfborder={0 0 1},backref=false,colorlinks=false]
 {hyperref}

\begin{document}

\title{The Development of Dominance Stripes and Orientation Maps in a Self-Organising
Visual Cortex Network (VICON)%
\thanks{This paper was submitted to Network on 6 November 1996. Paper reference
NET/79294/PAP. It was not accepted for publication, but it underpins
several subsequently published papers.%
}}

\author{S P Luttrell}

\maketitle
\noindent A self-organising neural network is presented that is based
on a rigorous Bayesian analysis of the information contained in individual
neural firing events. This leads to a visual cortex network (VICON)
that has many of the properties emerge when a mammalian visual cortex
is exposed to data arriving from two imaging sensors (i.e. the two
retinae), such as dominance stripes and orientation maps.

\section{Introduction}

The overall goal of this work is to automate as far as is possible
the processing of data from multiple sensors (data fusion), which
includes the automatic design of the architecture and functionality
of the network(s) that do the processing. In \cite{Luttrell1995c}
a novel approach to this automation problem was introduced, and the
purpose of this paper is to refine and extend the previously reported
results.

The problem of automating the design of a data fusion network has
many interesting special case solutions. In particular, the type of
self-organising neural network (in the mammalian visual cortex) that
processes the images arriving from a pair of retinae is one such special
case, where the number of sensors involved is just two. For a review
of visual cortex neural network models see \cite{Goodhill1992,Swindale1996}.

The basic idea is to use a soft encoder (i.e. its output is a distributed
code in which more than one, and possibly all, of the output neurons
is active) to transform the input vector (i.e. the input image) into
a posterior probability over various possible classes (i.e. alternative
possible interpretations of the input vector), and to optimise the
encoder so that this posterior probability is able to retain as much
information as possible about the input vector, as measured in the
minimum mean square reconstruction error (i.e. $L_{2}$ error) sense
\cite{Luttrell1995a,Luttrell1995b}.

In the special case where the optimisation is performed over the space
of all possible soft encoders, the optimum solution is a hard encoder
(i.e. it is a {}``winner-take-all'' network in which only one of
the output neurons is active) which is an optimal vector quantiser
(VQ), of the type described in \cite{Linde1980},\ for encoding the
input vector with minimum $L_{2}$ error. In the slightly less special
case where the space of possible soft encoders is restricted to include
only those whose output is deliberately damaged by the effects of
a noise process, this produces a different type of hard encoder which
is an optimal self-organising map (SOM) for encoding the input vector
with minimum $L_{2}$ error; this is very closely related to the well-known
Kohonen map \cite{Kohonen1984}, as was demonstrated in \cite{Luttrell1990}.

This paper will examine yet another special case, where the optimisation
is performed over a very special subspace of soft encoders, rather
than over all possible soft encoders. The behaviour of each soft encoder
is modelled by a set of posterior probabilities over various possible
classes. When a particular parametric form for these posterior probabilities
is chosen, a corresponding subspace of possible soft encoders is thus
automatically selected, which may be explored by varying the parameters.
The parametric form of the posterior probability that is used in this
paper is based on the so-called partitioned mixture distribution (PMD)
\cite{Luttrell1994b,Luttrell1994c}, which is a natural generalisation
of the standard mixture distribution to a high-dimensional input space.

This use of a PMD leads to a 2-layer visual cortex network (VICON),
where the components of the input vector are the output activities
of the input neurons, and the components of the PMD\ posterior probability
are the output activities of the output neurons. Various physically
realistic constraints are placed on the PMD optimisation (both on
the internal PMD structure, and on the type of training data that
is used), and these will be described in the text as they arise.

The layout of this paper is as follows. In section \ref{Sect:Theory}
all of the necessary theoretical machinery is developed, including
folded Markov chains, posterior probability models, derivatives of
the objective function, and receptive fields. In section \ref{Sect:DominanceOrientation}
the concepts of dominance stripes and orientation maps are explained,
both in the context of the elastic net model, and in the context of
theory presented in this paper. In section \ref{Sect:Simulations}
the results of computer simulations are presented, including both
1 and 2-dimensional retinae, single and pairs of retinae, both for
synthetic and natural training data. In appendix \ref{Sect:Optimal}
some explicit optimal solutions that minimise the objective function
are derived, including the periodicity property of some types of optimal
solution.

\section{Theory\label{Sect:Theory}}

This section covers all of the basic theoretical machinery that is
required to design and train a 2-layer VICON. In section \ref{Sect:FMC}
the theory of folded Markov chains (FMC) is summarised. In section
\ref{Sect:PostProbBasic} the basic idea of a posterior probability
model is introduced, and in section \ref{Sect:PostProbPMD} this is
developed into a full partitioned posterior probability model. In
section \ref{Sect:Derivative} the derivatives of the FMC objective
function are derived assuming a partitioned posterior probability
model, and in section \ref{Sect:Receptive} the influence of finite-sized
receptive fields on these derivatives is derived.

\subsection{Folded Markov Chain\label{Sect:FMC}}

\begin{figure}
\begin{centering}
\includegraphics[width=7cm]{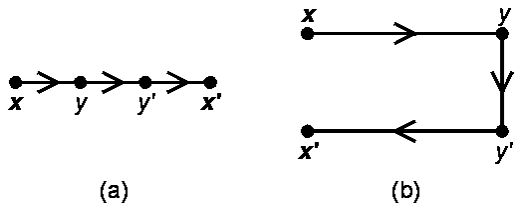}
\par\end{centering}

\begin{centering}
\caption{(a) A Markov chain of transitions $\mathbf{x\rightarrow}y\mathbf{\rightarrow}y^{\prime}\mathbf{\rightarrow x}^{\prime}$.
(b) The same diagram as (a), but folded.}

\par\end{centering}

\label{Fig:FMC}
\end{figure}

The basis of the entire theoretical treatment is a communication channel
model \cite{Luttrell1994a} in which an input vector $\mathbf{x}$
is encoded to produce a conditional probability $\Pr\left(y|\mathbf{x}\right)$
over code indices $y$, which is then transmitted along a distorting
communication channel to produce a conditional probability $\Pr\left(y^{\prime}|y\right)$
over distorted code indices $y^{\prime}$, which is finally decoded
to produce a conditional PDF $\Pr\left(\mathbf{x}^{\prime}|y^{\prime}\right)$
over reconstructions $\mathbf{x}^{\prime}$ of the original input
vector $\mathbf{x}$. The three steps in the sequence $\mathbf{x\rightarrow}y\mathbf{\rightarrow}y^{\prime}\mathbf{\rightarrow x}^{\prime}$
are modelled by the conditional probabilities $\Pr\left(y|\mathbf{x}\right)$,
$\Pr\left(y^{\prime}|y\right)$, and $\Pr\left(\mathbf{x}^{\prime}|y^{\prime}\right)$,
which describe a Markov chain of transitions, which is shown diagrammatically
in figure \ref{Fig:FMC}(a). $\Pr\left(\mathbf{x}^{\prime}|y\right)$
is completely determined from other defined quantities by using Bayes'
theorem in the form $\Pr\left(\mathbf{x}|y\right)=\frac{\Pr\left(\mathbf{x}\right)\Pr\left(y|\mathbf{x}\right)}{\int d\mathbf{x}^{\prime}\Pr\left(\mathbf{x}^{\prime}\right)\Pr\left(y|\mathbf{x}^{\prime}\right)}$.

Because $\mathbf{x}$ and $\mathbf{x}^{\prime}$ live in the same
vector space it is convenient to fold this diagram to produce figure
\ref{Fig:FMC}(b); this is called a folded Markov chain (FMC) \cite{Luttrell1994a}.
Figure \ref{Fig:FMC}(b) is directly related to a 2-layer unsupervised
neural network, where $\mathbf{x}$ and $\mathbf{x}^{\prime}$ represent
the activity pattern of the whole set of neurons in the input layer,
and $y$ and $y^{\prime}$ represent the location(s) of winning neuron(s)
in the ouput layer. The overall conditional PDF generated by an FMC
is $\Pr(\mathbf{x}^{\prime}|\mathbf{x})$, which is obtained by marginalising
$y$ and $y^{\prime}$ in the joint probability $\Pr\left(\mathbf{x}^{\prime},y^{\prime},y|\mathbf{x}\right)=\Pr\left(\mathbf{x}^{\prime}|y^{\prime}\right)\Pr\left(y^{\prime}|y\right)\Pr\left(y|\mathbf{x}\right)$.

Define a network objective function $D$ as \cite{Luttrell1994a}

\begin{equation}
D=\int d\mathbf{x}\, d\mathbf{x}^{\prime}\Pr\left(\mathbf{x}\right)\Pr(\mathbf{x}^{\prime}|\mathbf{x})\,\left\Vert \mathbf{x}-\mathbf{x}^{\prime}\right\Vert ^{2}\end{equation}
which measures the expected Euclidean (or $L_{2}$) reconstruction
error caused by feeding input vectors sampled from $\Pr\left(\mathbf{x}\right)$
into the FMC, where each $\mathbf{x}$ is returned as a PDF $\Pr(\mathbf{x}^{\prime}|\mathbf{x})$
of alternative reconstructions $\mathbf{x}^{\prime}$ of $\mathbf{x}$.
For simplicity, assume that the communication channel has been assumed
to be distortionless so that $\Pr\left(y^{\prime}|y\right)=\delta_{yy^{\prime}}$,
and that $y=1,2,\cdots,M$, then \begin{equation}
D=\sum_{y=1}^{M}\int d\mathbf{x}\, d\mathbf{x}^{\prime}\Pr\left(\mathbf{x}\right)\Pr\left(y|\mathbf{x}\right)\Pr\left(\mathbf{x}^{\prime}|y\right)\,\left\Vert \mathbf{x}-\mathbf{x}^{\prime}\right\Vert ^{2}\label{Eq:Objective}\end{equation}

An FMC is completely described by the form of its encoder $\Pr\left(y|\mathbf{x}\right)$
and the form of its reconstruction error $\left\Vert \mathbf{x}-\mathbf{x}^{\prime}\right\Vert ^{2}$.
The functional form of the encoder may be chosen arbitrarily, and
independently of the assumed Euclidean form of the reconstruction
error, so the FMC does \textit{not} correspond to a Gaussian mixture
distribution model in input space. This is a general result for FMCs
in which the functional forms of the encoder and the reconstruction
error may be independently chosen. It is only when these functional
forms are carefully chosen that a density model interpretation of
an FMC is possible (for instance a Euclidean reconstruction error
$\left\Vert \mathbf{x}-\mathbf{x}^{\prime}\right\Vert ^{2}$ must
be paired with an encoder $\Pr\left(y|\mathbf{x}\right)$ that describes
the posterior probability over class labels that would arise in a
Gaussian mixture distribution model).

The expression for $D$ given in equation \ref{Eq:Objective} may
be simplified to yield \cite{Luttrell1994a} (this readily generalises
to the case where $\Pr\left(y^{\prime}|y\right)\neq\delta_{yy^{\prime}}$
(i.e. the communication channel causes distortion)) \begin{equation}
D=2\int d\mathbf{x}\Pr\left(\mathbf{x}\right)\sum_{y=1}^{M}\Pr\left(y|\mathbf{x}\right)\left\Vert \mathbf{x}-\mathbf{x}^{\prime}\left(y\right)\right\Vert ^{2}\label{Eq:ObjectiveSimplified}\end{equation}
where $\mathbf{x}^{\prime}\left(y\right)$ is a reference vector defined
as \begin{equation}
\mathbf{x}^{\prime}\left(y\right)\equiv\int d\mathbf{x}\Pr\left(\mathbf{x}|y\right)\mathbf{x}\label{Eq:Reference}\end{equation}
If this definition of $\mathbf{x}^{\prime}\left(y\right)$ is not
used, and instead $D$ in equation \ref{Eq:ObjectiveSimplified} is
minimised with respect to $\mathbf{x}^{\prime}\left(y\right)$, then
the stationary solution is $\mathbf{x}^{\prime}\left(y\right)=\int d\mathbf{x}\Pr\left(\mathbf{x}|y\right)\mathbf{x}$,
which is consistent with the definition in equation \ref{Eq:Reference}.
In practice, it is better to determine the stationary $\mathbf{x}^{\prime}\left(y\right)$
by following the gradient $\frac{\partial D}{\partial\mathbf{x}^{\prime}\left(y\right)}$
than to use the explicit expression $\int d\mathbf{x}\Pr\left(\mathbf{x}|y\right)\mathbf{x}$
for the stationary point, because $\frac{\partial D}{\partial\mathbf{x}^{\prime}\left(y\right)}$
is cheap to evaluate whereas $\int d\mathbf{x}\Pr\left(\mathbf{x}|y\right)\mathbf{x}$
is expensive to evaluate. In effect, the $\frac{\partial D}{\partial\mathbf{x}^{\prime}\left(y\right)}$
approach is an example of on-line training, whereas the $\int d\mathbf{x}\Pr\left(\mathbf{x}|y\right)\mathbf{x}$
approach is the corresponding example of batch training, and the on-line
and batch approaches each have their own areas where they are best
used.

In equation \ref{Eq:ObjectiveSimplified} $\Pr\left(y|\mathbf{x}\right)$
is a {}``recognition model'' (i.e. it takes an input vector and
recognises by assigning to it a posterior probability over class labels)
and $\mathbf{x}^{\prime}\left(y\right)$ is the corresponding {}``generative
model'' (i.e. it takes a class label and generates a corresponding
vector in input space). This is a simpler type of generative model
than appeared in the original expression for $D$ in equation \ref{Eq:Objective},
where the generative model is $\Pr\left(\mathbf{x}^{\prime}|y\right)$,
which generates a whole distribution of possible vectors in input
space, rather than just a single vector which is the centroid of $\Pr\left(\mathbf{x}^{\prime}|y\right)$.
The transformation of the FMC from one that uses the PDF $\Pr\left(\mathbf{x}^{\prime}|y\right)$
into one that uses the reference vector $\mathbf{x}^{\prime}\left(y\right)$
is not possible in general; it was made possible here by choosing
to use a Euclidean reconstruction error in $D$. In general, an FMC
reconstruction is a distribution over alternative inputs, rather than
a single representative input, as might be used in decision theory,
for instance.

\begin{figure}
\begin{centering}
\includegraphics[width=7cm]{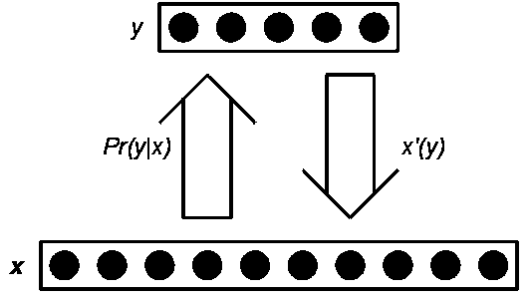}
\par\end{centering}

\begin{centering}
\caption{A neural network representation of a folded Markov chain.}

\par\end{centering}

\label{Fig:FMCNN}
\end{figure}

The operation of the various terms in the expression for $D$ in equation
\ref{Eq:ObjectiveSimplified} is shown in figure \ref{Fig:FMCNN},
which is rotated through 90${^{\circ}}$ anticlockwise with respect
to the corresponding diagram in figure \ref{Fig:FMC}, and also for
simplicity $y^{\prime}=y$ because $\Pr\left(y^{\prime}|y\right)=\delta_{yy^{\prime}}$
was assumed above. %
 When $D$ is minimised with respect to the choice of encoder $\Pr\left(y|\mathbf{x}\right)$
and reconstruction vector $\mathbf{x}^{\prime}\left(y\right)$ it
yields a standard minimum mean square error vector quantiser (VQ)
with $M$ code indices \cite{Linde1980}, and if $\Pr\left(y^{\prime}|y\right)\neq\delta_{yy^{\prime}}$
then the VQ produces code indices that carry information in such a
way that it is maximally robust with respect to the damaging effects
of communication channel distortion modelled by $\Pr\left(y^{\prime}|y\right)$
\cite{Luttrell1990}. This latter type of VQ can be shown to be approximately
equivalent to a self-organising map (SOM) of the type introduced by
Kohonen \cite{Kohonen1984}.

\subsection{Basic Posterior Probability (Single Recognition Model)\label{Sect:PostProbBasic}}

The minimisation procedure that leads to a VQ-like optimum assumed
that the entire space of posterior probability functions $\Pr\left(y|\mathbf{x}\right)$
was available to be searched. In the neural network interpretation,
$\Pr\left(y|\mathbf{x}\right)$ models the probability that neuron
$y$ fires first (this encompasses both the case of a soft encoder
where more than one neuron can potentially fire first, and the case
of a hard encoder where only one neuron can potentially fire first;
this is the winner-take-all case), which depends on the detailed underlying
dynamics of how all of the neurons interact with each other. Because
these neural dynamics are not arbitrary (e.g. they are constrained
to be a physically realisable process), it constrains the space of
possible posterior probabilities $\Pr\left(y|\mathbf{x}\right)$ that
is available to the neural network. $\Pr\left(y|\mathbf{x}\right)$
may then be modelled by the functional form \begin{equation}
\Pr\left(y|\mathbf{x}\right)\equiv\frac{Q\left(\mathbf{x|}y\right)}{\sum_{y^{\prime}=1}^{M}Q\left(\mathbf{x|}y^{\prime}\right)}\label{Eq:Posterior}\end{equation}
 where $Q\left(\mathbf{x|}y\right)$ is the raw {}``response function''
of neuron $y$.

$Q\left(\mathbf{x|}y\right)$ may be interpreted as the raw firing
rate of neuron $y$, and $\Pr\left(y|\mathbf{x}\right)$ is then the
probability that neuron $y$ fires first out of all of the $M$ competing
neurons. This functional form makes it clear that there is a type
of lateral inhibition occurring between $\Pr\left(y_{1}|\mathbf{x}\right)$
and $\Pr\left(y_{2}|\mathbf{x}\right)$ (for $y_{1}\neq y_{2}$),
because if the raw firing rate $Q\left(\mathbf{x|}y_{1}\right)$ is
\textit{increased} so that $\Pr\left(y_{1}|\mathbf{x}\right)$ increases,
nevertheless the denominator $\sum_{y^{\prime}=1}^{M}Q\left(\mathbf{x|}y^{\prime}\right)$
ensures that $\Pr\left(y_{2}|\mathbf{x}\right)$ \textit{decreases}
(for $y_{1}\neq y_{2}$); i.e. the $Q\left(\mathbf{x|}y\right)$ do
not exhibit lateral inhibition, but the $\Pr\left(y_{1}|\mathbf{x}\right)$
do exhibit lateral inhibition.

The raw receptive field of a neuron depends on the form of $Q\left(\mathbf{x|}y\right)$.
Thus if the functional form of $Q\left(\mathbf{x|}y\right)$ depends
only on a subset $\mathbf{\tilde{x}}\left(y\right)$ of components
of $\mathbf{x}$, then $\mathbf{\tilde{x}}\left(y\right)$ is the
raw receptive field of neuron $y$. However, this is \textit{not}
the same as the the receptive field that is effective in producing
the first firing event, because $\Pr\left(y|\mathbf{x}\right)$ depends
on all of the $\mathbf{\tilde{x}}\left(y^{\prime}\right)$ (for $y^{\prime}=1,\cdots,M$)
as shown in equation \ref{Eq:Posterior}.

The effect of the distortion $\Pr\left(y\mathbf{|}y\mathbf{^{\prime}}\right)$
process, as modelled by $\Pr\left(y\mathbf{|}y\mathbf{^{\prime}}\right)$,
is to alter at the last minute, as it were, the probability that each
neuron fires first. Thus the posterior probability is modified as
follows \begin{equation}
\Pr\left(y|\mathbf{x}\right)\rightarrow\sum_{y^{\prime}=1}^{M}\Pr\left(y\mathbf{|}y\mathbf{^{\prime}}\right)\Pr\left(y^{\prime}|\mathbf{x}\right)\label{Eq:Leakage}\end{equation}
 where the matrix element $\Pr\left(y\mathbf{|}y\mathbf{^{\prime}}\right)$
leaks posterior probability from neuron $y^{\prime}$ onto neuron
$y$. Such cross-talk amongst the neurons exists independently of
the lateral inhibition effect produced by the denominator term $\sum_{y^{\prime}=1}^{M}Q\left(\mathbf{x|}y^{\prime}\right)$
in equation \ref{Eq:Posterior}.

The VQ and SOM results (see \cite{Linde1980,Kohonen1984})\ may be
obtained as special cases of raw neuron firing rates $Q\left(\mathbf{x|}y\right)$,
where one neuron's firing rate is much larger than the other $M-1$
neurons' firing rates (i.e. there is effectively only one neuron that
can fire, so it is the winner-take-all).

\subsection{Partitioned Posterior Probability (Multiple Recognition Models)\label{Sect:PostProbPMD}}

The form of the posterior probability $\Pr\left(y|\mathbf{x}\right)$
introduced in equation \ref{Eq:Posterior} is unsuitable for networks
with a large number of neurons $M$, because the lateral inhibition
is \textit{global} rather than \textit{local}. This can readily be
inferred because the denominator term $\sum_{y^{\prime}=1}^{M}Q\left(\mathbf{x|}y^{\prime}\right)$
in equation \ref{Eq:Posterior} computes a quantity that is the sum
over \textit{all} of the raw neuron firing rates.

This problem can be amended by defining a \textit{localised} posterior
probability $\Pr\left(y|\mathbf{x;}y^{\prime}\right)$ as

\begin{equation}
\Pr\left(y|\mathbf{x;}y^{\prime}\right)\equiv\frac{Q\left(\mathbf{x|}y\right)\delta_{y\in\mathcal{N}\left(y^{\prime}\right)}}{\sum_{y^{\prime\prime}\in\mathcal{N}\left(y^{\prime}\right)}Q\left(\mathbf{x|}y^{\prime\prime}\right)}\label{Eq:PosteriorLocal}\end{equation}
 where $\mathcal{N}\left(y^{\prime}\right)$ is the \textit{local}
neighbourhood of neuron $y^{\prime}$, which is assumed to contain
at least neuron $y^{\prime}$, and $\delta_{y\in\mathcal{N}\left(y^{\prime}\right)}$
is a Kronecker delta that constrains $y$ to lie in the neighbourhood
$\mathcal{N}\left(y^{\prime}\right)$. If $\mathcal{N}\left(y^{\prime}\right)$
contains all $M$ neurons then $\Pr\left(y|\mathbf{x;}y^{\prime}\right)$
reduces to $\Pr\left(y|\mathbf{x}\right)$ as previously defined in
equation \ref{Eq:Posterior}. $\Pr\left(y|\mathbf{x;}y^{\prime}\right)$
has the required normalisation property that $\sum_{y=1}^{M}\Pr\left(y|\mathbf{x;}y^{\prime}\right)=1$
for all $y^{\prime}$. Because $y^{\prime}$ can take $M$ possible
values, there are $M$ complete localised posterior probability functions
$\Pr\left(y|\mathbf{x;}y^{\prime}\right)$. In effect, the neural
network is split up into $M$ overlapping subnetworks (these subnetworks
overlap where $\mathcal{N}\left(y_{1}\right)\cap\mathcal{N}\left(y_{2}\right)\neq\emptyset$
for $y_{1}\neq y_{2}$), each of which computes its own posterior
probability function; note that any overlap between a pair of subnetworks
causes the corresponding $\Pr\left(y|\mathbf{x;}y^{\prime}\right)$
to be mutually dependent.

It is not always convenient to use a neural network model in which
there are $M$ separate posterior probability models $\Pr\left(y|\mathbf{x;}y^{\prime}\right)$.
However, these $M$ localised posterior probability functions $\Pr\left(y|\mathbf{x;}y^{\prime}\right)$
(for the $M$ different choices of $y^{\prime}$)\ may be averaged
together to produce a \textit{single} posterior probability function.
Thus define $\Pr\left(y|\mathbf{x}\right)$ as

\begin{eqnarray}
\Pr\left(y|\mathbf{x}\right) & \equiv & \frac{1}{M}\sum_{y^{\prime}\in\mathcal{N}^{-1}\left(y\right)}\Pr\left(y|\mathbf{x;}y^{\prime}\right)\nonumber \\
 & = & \frac{1}{M}\, Q\left(\mathbf{x|}y\right)\sum_{y^{\prime}\in\mathcal{N}^{-1}\left(y\right)}\frac{1}{\sum_{y^{\prime\prime}\in\mathcal{N}\left(y^{\prime}\right)}Q\left(\mathbf{x|}y^{\prime\prime}\right)}\label{Eq:PosteriorPMD}\end{eqnarray}
 where $\mathcal{N}^{-1}\left(y\right)$ is the inverse neighbourhood
of neuron $y$ defined as $\mathcal{N}^{-1}\left(y\right)\equiv\left\{ y^{\prime}|y\in\mathcal{N}\left(y^{\prime}\right)\right\} $.
This definition has all of the properties of a posterior probability
function, including the normalisation property $\sum_{y=1}^{M}\Pr\left(y|\mathbf{x}\right)=1$
(which may be derived by swapping the order of summations using the
result $\sum_{y=1}^{M}\sum_{y^{\prime}\in\mathcal{N}^{-1}\left(y\right)}\left(\cdots\right)=\sum_{y^{\prime}=1}^{M}\sum_{y\in\mathcal{N}\left(y^{\prime}\right)}\left(\cdots\right)$).
The form of the posterior probability given in equation \ref{Eq:PosteriorPMD},
in which $M$ individual posterior probabilities $\Pr\left(y|\mathbf{x;}y^{\prime}\right)$
are averaged together, can be rigorously justified from a Bayesian
point of view (see appendix \ref{Sect:BayesPMD}). The averaging process
produces the posterior probability that should be used when there
are $M$ contributing models (as specified by the $\Pr\left(y|\mathbf{x;}y^{\prime}\right)$
for $y^{\prime}=1,2,\cdots,M$) that\ have equal prior weight. The
average over the $M$ models then simply marginalises over an unobserved
degree of freedom (the model index $y^{\prime}$).

If this localised definition of $\Pr\left(y|\mathbf{x}\right)$ given
in equation \ref{Eq:PosteriorPMD} is compared with the global definition
given in equation \ref{Eq:Posterior} it is seen that the normalisation
factor has been modified thus \begin{equation}
\frac{1}{\sum_{y^{\prime}=1}^{M}Q\left(\mathbf{x|}y^{\prime}\right)}\rightarrow\frac{1}{M}\sum_{y^{\prime}\in\mathcal{N}^{-1}\left(y\right)}\frac{1}{\sum_{y^{\prime\prime}\in\mathcal{N}\left(y^{\prime}\right)}Q\left(\mathbf{x|}y^{\prime\prime}\right)}\label{Eq:NormalisationChange}\end{equation}
 which alters its lateral inhibition properties. $\frac{1}{\sum_{y^{\prime\prime}\in\mathcal{N}\left(y^{\prime}\right)}Q\left(\mathbf{x|}y^{\prime\prime}\right)}$
is the lateral inhibition factor that derives from the neighbourhood
of neuron $y^{\prime}$, which gives rise to a contribution to the
lateral inhibition factor for all neurons $y$ in the neighbourhood
of $y^{\prime}$ via the average $\frac{1}{M}\sum_{y^{\prime}\in\mathcal{N}^{-1}\left(y\right)}\left(\cdots\right)$.
Thus the overall lateral inhibition factor acting on neuron $y$ is
derived \textit{locally} from those neurons $y^{\prime\prime}$ that
lie in the set $\mathcal{N}\left(\mathcal{N}^{-1}\left(y\right)\right)$.
The posterior probability model defined in equation \ref{Eq:PosteriorPMD}
has been used before in the context of partitioned mixture distributions
(PMDs), where multiple mixture distribution models are simultaneously
optimised \cite{Luttrell1994b,Luttrell1994c}.

\begin{figure}
\begin{centering}
\includegraphics[width=7cm]{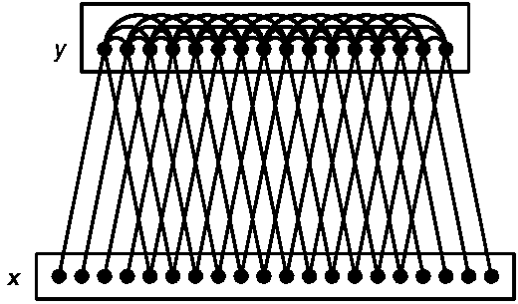}
\par\end{centering}

\caption{A partitioned mixture distribution (PMD) neural network.}

\label{Fig:PMD}
\end{figure}

Figure \ref{Fig:PMD} shows the structure of the neural network corresponding
to the PMD posterior probability in equation \ref{Eq:PosteriorPMD}.
 Each output neuron has a raw receptive field of input neurons (which
contains 5 input neurons in the example shown), and is also laterally
inhibited by its neighbouring output neurons (the size of a neuron
neighbourhood is 3 neurons to either side in the example shown). Note
that the input-output links in figure \ref{Fig:PMD} do not imply
that the raw neuron firing rates $Q\left(\mathbf{x|}y\right)$ can
be computed by using simple weighted connections; they are drawn merely
to indicate the set of input neurons that influences the raw firing
rate of each output neuron. Similarly, the output-output links in
figure \ref{Fig:PMD} are drawn to indicate the sizes of the output
neuron neighbourhoods; the details of how lateral inhibition modifies
the raw firing rates $Q\left(\mathbf{x|}y\right)$ of the output neurons
to produce the probability $\Pr\left(y|\mathbf{x}\right)$ that neuron
$y$ fires first is given in equation \ref{Eq:PosteriorPMD}.

For completeness, the PMD objective function in equation \ref{Eq:ObjectiveSimplified}
may now be written out in full using the expression for the PMD posterior
probability in equation \ref{Eq:PosteriorPMD} to yield (where the
effects of leakage have been included, as defined in equation \ref{Eq:Leakage})
\begin{eqnarray}
D & = & \frac{2}{M}\int d\mathbf{x}\Pr\left(\mathbf{x}\right)\sum_{y=1}^{M}\sum_{y^{\prime}=1}^{M}\Pr\left(y\mathbf{|}y\mathbf{^{\prime}}\right)\nonumber \\
 &  & \times\, Q\left(\mathbf{x|}y^{\prime}\right)\sum_{y^{\prime\prime}\in\mathcal{N}^{-1}\left(y^{\prime}\right)}\frac{1}{\sum_{y^{\prime\prime\prime}\in\mathcal{N}\left(y^{\prime\prime}\right)}Q\left(\mathbf{x|}y^{\prime\prime\prime}\right)}\left\Vert \mathbf{x}-\mathbf{x}^{\prime}\left(y\right)\right\Vert ^{2}\label{Eq:ObjectiveComplete}\end{eqnarray}
 This is the objective function that will be used to characterise
to performance of the neural networks in all of the computer simulations.

\subsection{Derivatives of the Objective Function\label{Sect:Derivative}}

In order to minimise the PMD objective function in equation \ref{Eq:ObjectiveComplete}
its derivatives must be calculated. First of all, define some convenient
notation \cite{Luttrell1995b}

\begin{equation}
\begin{array}{ll}
L_{y,y^{\prime}}\equiv\Pr\left(y\mathbf{^{\prime}|}y\right) & P_{y,y^{\prime}}\equiv\Pr\left(y\mathbf{^{\prime}|x;}y\right)\equiv\frac{Q\left(\mathbf{x|}y^{\prime}\right)\delta_{y^{\prime}\in\mathcal{N}\left(y\right)}}{\sum_{y^{\prime\prime}\in\mathcal{N}\left(y\right)}Q\left(\mathbf{x|}y^{\prime\prime}\right)}\\
p_{y}\equiv\sum_{y^{\prime}\in\mathcal{N}^{-1}\left(y\right)}P_{y^{\prime},y} & \left(L^{T}p\right)_{y}\equiv\sum_{y^{\prime}\in\mathcal{L}^{-1}\left(y\right)}L_{y^{\prime}\mathbf{,}y}p_{y^{\prime}}\\
e_{y}\equiv\left\Vert \mathbf{x}-\mathbf{x}^{\prime}\left(y\right)\right\Vert ^{2} & \left(Le\right)_{y}\equiv\sum_{y^{\prime}\in\mathcal{L}\left(y\right)}L_{y,y^{\prime}}e_{y^{\prime}}\\
\left(PLe\right)_{y}\equiv\sum_{y^{\prime}\in\mathcal{N}\left(y\right)}P_{y,y^{\prime}}\left(Le\right)_{y^{\prime}} & \left(P^{T}PLe\right)_{y}\equiv\sum_{y^{\prime}\in\mathcal{N}^{-1}\left(y\right)}P_{y\mathbf{^{\prime},}y}\left(PLe\right)_{y^{\prime}}\end{array}\label{Eq:Notation}\end{equation}
 where $\mathcal{L}\left(y\right)$ denotes the leakage neighbourhood
of neuron $y$, which is the set of neurons that have posterior probability
leaked onto them by neuron $y$, and the inverse leakage neighbourhood
$\mathcal{L}^{-1}\left(y\right)$ is defined as $\mathcal{L}^{-1}\left(y\right)\equiv\left\{ y^{\prime}|y\in\mathcal{L}\left(y^{\prime}\right)\right\} $.
Assume that the raw neuron firing rates may be modelled using a sigmoid
function \begin{equation}
Q\left(\mathbf{x}|y\right)=\frac{1}{1+\exp\left(-\mathbf{w}\left(y\right)\cdot\mathbf{x}-b\left(y\right)\right)}\label{Eq:Sigmoid}\end{equation}
 whence the derivatives may be obtained in the form \cite{Luttrell1995b}
\begin{eqnarray}
\frac{\partial D}{\partial\mathbf{x}^{\prime}\left(y\right)} & = & -\frac{4}{M}\int d\mathbf{x}\Pr\left(\mathbf{x}\right)\left(L^{T}p\right)_{y}\left(\mathbf{x}-\mathbf{x}^{\prime}\left(y\right)\right)\nonumber \\
\frac{\partial D}{\partial\left(\begin{array}{l}
b\left(y\right)\\
\mathbf{w}\left(y\right)\end{array}\right)} & = & \frac{2}{M}\int d\mathbf{x}\Pr\left(\mathbf{x}\right)\left[\begin{array}{c}
\left(p_{y}\left(Le\right)_{y}-(P^{T}PLe)_{y}\right)\\
\times\left(1-Q\left(\mathbf{x}|y\right)\right)\left(\begin{array}{l}
1\\
\mathbf{x}\end{array}\right)\end{array}\right]\label{Eq:ObjectiveDerivative}\end{eqnarray}
 where the two derivatives $\frac{\partial D}{\partial b\left(y\right)}$
and $\frac{\partial D}{\partial\mathbf{w}\left(y\right)}$ have been
written together for compactness.

\subsection{Receptive Fields\label{Sect:Receptive}}

The raw firing rate $Q\left(\mathbf{x}|y\right)$ of neuron $y$ depends
only on a subset $\mathbf{\tilde{x}}\left(y\right)$ of components
of $\mathbf{x}$; $\mathbf{\tilde{x}}\left(y\right)$ is thus the
\textit{raw} receptive field of neuron $y$. However, the posterior
probability $\Pr\left(y|\mathbf{x}\right)$ that neuron $y$ fires
first is derived from $Q\left(\mathbf{x}|y\right)$ by weighting it
with a lateral inhibition factor that depends on the raw firing rates
of all neurons in $\mathcal{N}\left(\mathcal{N}^{-1}\left(y\right)\right)$,
as seen in equation \ref{Eq:PosteriorPMD}, so the \textit{overall}
receptive field of a neuron is rather broader than its raw receptive
field. The effect of leakage, as defined in equation \ref{Eq:Leakage},
is to broaden the overall receptive field further still. The optimal
reference vector $\mathbf{x}^{\prime}\left(y\right)$ has non-trivial
structure only within this overall receptive field, so inside the
overall receptive field the components of $\mathbf{x}^{\prime}\left(y\right)$
must be subjected to an optimisation procedure to discover their optimal
form, whereas outside the overall receptive field the components of
$\mathbf{x}^{\prime}\left(y\right)$ may be set to be the average
values of the corresponding components of the training vectors $\mathbf{x}$
(see the definition of $\mathbf{x}^{\prime}\left(y\right)$ in equation
\ref{Eq:Reference}, which reduces to $\mathbf{x}^{\prime}\left(y\right)=\int d\mathbf{x}\Pr\left(\mathbf{x}\right)\mathbf{x}$
for those components of $\mathbf{x}^{\prime}\left(y\right)$ that
lie outside the overall receptive field of neuron $y$).

In the simulations that will be presented here a suboptimal approach
is used, where only those components of $\mathbf{x}^{\prime}\left(y\right)$
that lie inside the \textit{raw} receptive field are optimised; this
produces a least upper bound on the value of the objective function
that would have been obtained if a full optimisation had been used.
Also, it is assumed that the input data has been prepared in such
a way that each component is zero mean. This is not actually a restriction,
because the objective function is invariant with respect to adding
a different constant to each component of $\mathbf{x}$, because it
is a function of the difference $\mathbf{x}-\mathbf{x}^{\prime}$.
In this suboptimal approach, and with the zero mean assumption, the
components of $\mathbf{x}^{\prime}\left(y\right)$ that lie outside
the \textit{raw} receptive field of neuron $y$ will be set to zero.

The fact that the components of $\mathbf{x}^{\prime}\left(y\right)$
that lie outside the \textit{raw} receptive field of neuron $y$ are
zero may be used to simplify the evaluation of the various terms $\frac{\partial D}{\partial b\left(y\right)}$
and $\frac{\partial D}{\partial\mathbf{w}\left(y\right)}$ in equation
\ref{Eq:ObjectiveDerivative}. Thus evaluate $p_{y}\left(Le\right)_{y}-(P^{T}PLe)_{y}$
by expanding $e_{y}$ as \begin{eqnarray}
e_{y} & = & \left\Vert \mathbf{x}\right\Vert ^{2}-2\mathbf{x}\cdot\mathbf{x}^{\prime}\left(y\right)+\left\Vert \mathbf{x}^{\prime}\left(y\right)\right\Vert ^{2}\nonumber \\
 & = & \left\Vert \mathbf{x}\right\Vert ^{2}+\mathbf{x}^{\prime}\left(y\right)\cdot\left(\mathbf{x}^{\prime}\left(y\right)-2\mathbf{x}\right)\label{Eq:ExpandEy}\end{eqnarray}
 which is a sum of a constant (i.e. does not depend on $y$) term
$\left\Vert \mathbf{x}\right\Vert ^{2}$ and a term $\mathbf{x}^{\prime}\left(y\right)\cdot\left(\mathbf{x}^{\prime}\left(y\right)-2\mathbf{x}\right)$
that does depend on $y$. What happens to the constant term when it
is substituted into $p_{y}\left(Le\right)_{y}-(P^{T}PLe)_{y}$? \begin{eqnarray}
p_{y}\left(Le\right)_{y}-(P^{T}PLe)_{y} & \rightarrow & p_{y}\left(L\cdot\mathbf{1}\right)_{y}-(P^{T}PL\cdot\mathbf{1})_{y}\nonumber \\
 & = & p_{y}\mathbf{1}_{y}-(P^{T}P\cdot\mathbf{1})_{y}\nonumber \\
 & = & p_{y}-p_{y}\nonumber \\
 & = & 0\label{Eq:KillConstant}\end{eqnarray}
 It cancels out, so $e_{y}$ might as well be replaced as follows
in $p_{y}\left(Le\right)_{y}-(P^{T}PLe)_{y}$ \begin{equation}
e_{y}\rightarrow\mathbf{x}^{\prime}\left(y\right)\cdot\left(\mathbf{x}^{\prime}\left(y\right)-2\mathbf{x}\right)\label{Eq:Projection}\end{equation}
 Because the components of $\mathbf{x}^{\prime}\left(y\right)$ that
lie outside the \textit{raw} receptive field of neuron $y$ are set
to zero, the $\mathbf{x}^{\prime}\left(y\right)\cdot\left(\cdots\right)$
operation effectively projects out any components of $\left(\cdots\right)$
that happen to lie outside this raw receptive field. This means that
the only components of $\mathbf{x}$ in equation \ref{Eq:Projection}
that survive are those that lie inside the raw receptive field, so
effectively $e_{y}$ depends only on quantities that lie inside the
raw receptive field of neuron $y$. Note that a full optimisation
of $\mathbf{x}^{\prime}\left(y\right)$, in which all components that
lie inside the \textit{overall} receptive field of neuron $y$ are
optimised, would produce a different result.

\section{Dominance Stripes and Orientation Maps\label{Sect:DominanceOrientation}}

The purpose of this section is to discuss the two phenomena of dominance
stripes and orientation maps. In section \ref{Sect:Elastic} a brief
review of the popular elastic net model of dominance stripes is presented,
and in section \ref{Sect:Informal} an informal derivation of the
origin of both dominance stripes and orientation maps is given.

\subsection{Review of Dominance Stripes Using the Elastic Net Model\label{Sect:Elastic}}

\begin{figure}
\begin{centering}
\includegraphics[width=7cm]{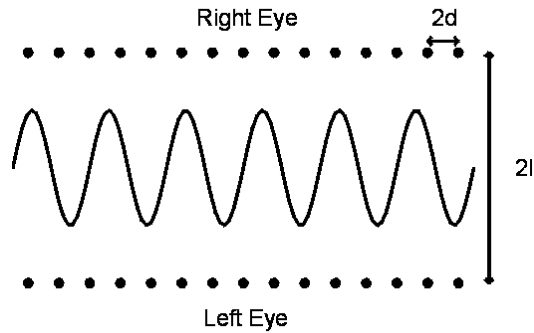}
\par\end{centering}

\caption{An elastic net oscillating back and forth in ocularity between a pair
of retinae.}

\label{Fig:Elastic}
\end{figure}

The results that will be presented here are, broadly speaking, equivalent
to the way in which ocular dominance stripes are obtained in the elastic
net model (as reviewed in \cite{Goodhill1992,Swindale1996}) as applied
to a pair of retinae. The essential features of this type of model
of ocular dominance are shown in figure \ref{Fig:Elastic} (which
is copied from \cite{Goodhill1992}). %
 The left and right retinae are represented as 1-dimensional lines
of units at the top and bottom of the diagram. The horizontal dimension
represents distance across a retina, and the vertical dimension represents
the ocularity degree of freedom. The distance between any two retinal
units, either within or between retinae, represents the correlation
between those two units \cite{Goodhill1992}. Thus the ratio $\frac{l}{d}$
determines the relative strength of the inter-retinal and intra-retinal
correlations. The elastic net is represented by the line oscillating
back and forth between the retinae. The net effect of the elastic
net algorithm is to encourage the elastic net to pass as close as
possible (in a well-defined sense) to all of the retinal units, and
also to minimise its total length. These are conflicting requirements,
and the oscillatory solution shown in figure \ref{Fig:Elastic} is
typical of an optimal elastic net configuration, which thus predicts
an oscillatory pattern of ocular dominance (i.e. which corresponds
to dominance stripes in the case of 2-dimensional retinae).

This type of model inevitably leads to dominance stripe formation,
because the elastic net model separates the input components into
two clusters (see figure \ref{Fig:Elastic}) according to whether
they belong to the left or right retina. In effect the output layer
of the network is explicitly told which retina an input component
belongs to, and this fact is expressed by the position of the component
along the ocularity dimension. The goal in this paper is to construct
a more natural model of dominance stripe formation, in which the ocularity
dimension is revealed by a process of self-organisation, rather than
being hard-wired into the model. Thus, the visual cortex model that
is presented in this paper will \textit{not} explicitly label the
input pixels as belonging to the right or left retina (as they are
in figure \ref{Fig:Elastic}), but will have to deduce their left/right
retina membership from the properties of the training set instead.

\subsection{Informal Derivation of Dominance Stripes and Orientation Maps\label{Sect:Informal}}

The purpose of this section is to present a simple picture that makes
it clear what types of behaviour should be expected from neural network
that minimises the objective function in equation \ref{Eq:ObjectiveComplete}.

\subsubsection{Neural Network Model}

It is assumed that each of the output neurons has only a limited receptive
field of input neurons within each of the two retinae. In effect,
this is a hand-crafted version of a {}``wire length'' constraint,
which ensures that the total length of the input-to-output connections
is limited. In the context of the elastic net model this corresponds
to the limited range of interaction between retinal units (the input)\ and
elastic net units (the output). Also, it is assumed that sigmoidal
neurons with local probability leakage are used, which generates an
effect that is analogous to the elastic tension in the elastic net
model, because it encourages neighbouring neurons to adopt similar
parameter values.

\begin{figure}
\begin{centering}
\includegraphics[width=7cm]{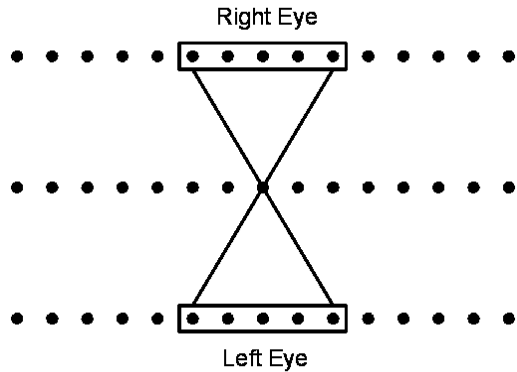}
\par\end{centering}

\caption{Neural network model with a limited receptive field.}

\label{Fig:Receptive}
\end{figure}

This model is drawn in figure \ref{Fig:Receptive} in an analogous
way to the elastic net model in figure \ref{Fig:Elastic}. %
 In this model the ocularity dimension is not explicitly present,
and the elasticity (of the elastic net) is replaced by the probability
leakage mechanism that enables neighbouring output neurons to communicate
with each other. The separation of input neurons into left and right
retinae in figure \ref{Fig:Receptive} is made only for comparison
between figure \ref{Fig:Receptive} and the elastic net model in figure
\ref{Fig:Elastic}. When the left and right receptive fields are presented
to the output neuron, all information about which retina the various
input neurons belong to has been discarded; all input neurons within
the left and right receptive fields are treated on an equal basis.
The ocularity dimension will emerge by a process of self-organisation
driven by the statistical properties of the images received by the
left and right retinae.

\subsubsection{Very Low Resolution Input Images}

The simplest situation is when there are two retinae (as in the above
elastic net model), each of which senses independently a featureless
scene, i.e. all the units in a retina sense the same brightness value,
but the two brightnesses that the left and right retinae sense are
independent of each other. This situation would arise if the images
projected onto the two retinae were very low resolution, so all spatial
detail is lost. This limits the input data to lying in a 2-dimensional
space $R^{2}$. If these two featureless input images (i.e. left and
right retinae) are then normalised so that the sum of left and right
retina brightness is constrained to be constant, then the input data
is projected down onto a 1-dimensional space $R^{1}$, which effectively
becomes the ocularity dimension. If each of the $M$ output neurons
had an infinite-sized receptive field, then the optimal network would
be the one in which the $M\,$neurons cooperate to give the best soft
encoding of $R^{1}$.

However, because of the limited receptive field size and output neuron
neighbourhood size, the neurons can at best co-operate together a
few at a time (this also depends on the size of the leakage neighbourhood).
If the network properties are translation invariant this leads to
an optimal network whose properties fluctuate periodically across
the network (see appendix \ref{Sect:Optimal}), where each period
typically contains a complete repertoire of the computing machinery
that is needed to process the contents of a receptive field; this
effect is called completeness, and it is a characteristic emergent
property of this type of neural network.

The only unexplained step in this argument is the use of a normalisation
procedure on the input. However, if the input to this network is the
PMD posterior probability computed by the output layer of another
such network, then there is already such a normalisation effect induced
by the lateral inhibition within the PMD posterior probability. For
featureless input images, this lateral inhibition effect causes precisely
the type of normalisation that is used above (i.e. left plus right
retina brightness is constant) to occur naturally.

\begin{figure}
\begin{centering}
\includegraphics[width=7cm]{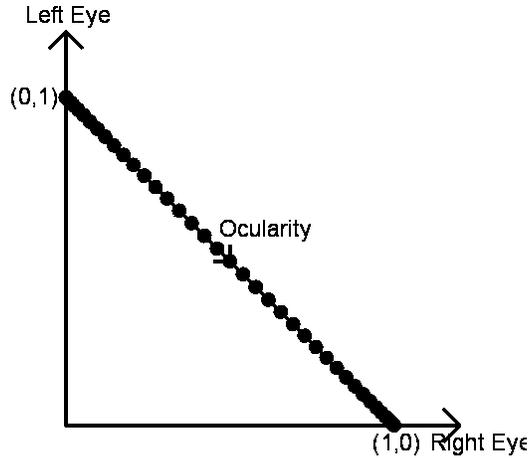}
\par\end{centering}

\caption{Typical neural reference vectors for very low resolution input images.}

\label{Fig:Ocular1}
\end{figure}

These results are summarised in figure \ref{Fig:Ocular1} %
 where the ocularity dimension runs from $\left(0,1\right)$ to $\left(1,0\right)$,
and a typical set of neural reference vectors is shown. The oscillation
of these reference vectors back and forth along the ocularity dimension
corresponds to the oscillations of the elastic net that are represented
in figure \ref{Fig:Elastic}.

\subsubsection{Low Resolution Input Images}

A natural generalisation of the above is to the case of not-quite-featureless
input images. This could be brought about by gradually increasing
the resolution of the input images until it is sufficient to reveal
spatial detail on a size scale equal to the receptive field size.
Instead of seeing a featureless input, each neuron would then see
a brightness gradient within its receptive field. This could be interpreted
by considering the low order terms of a Taylor expansion of the input
image about a point at the centre of the neuron's receptive field:
the zeroth term is local average brightness (which lives on a 1-dimensional
line $R^{1}$), and the two first order terms are the local brightness
gradient (which lives in a 2-dimensional space $R^{2}$). When normalisation
is applied this reduces the space in which the two images live to
$R^{1}\times R^{2}\times R^{2}$ ($R^{1}$ from the zeroth order Taylor
term with normalisation taken into account, $R^{2}$ from the first
order Taylor terms, counted twice to deal with each retina).

The $R^{1}$ from the zeroth order Taylor term gives rise to ocular
dominance stripes, which thus causes the left and right retinae to
map to different stripe-shaped regions of the output layer. The remaining
$R^{2}\times R^{2}$ then naturally splits into two contributions
(left retina and right retina), each of which maps to the appropriate
stripe. If the stripes did not separate the left and right retinae,
then the $R^{2}\times R^{2}$ could not be split apart in this simple
manner. Finally, since each ocular dominance stripe occupies a 2-dimensional
region of the output layer, a direct mapping of the corresponding
$R^{2}$ (which carries local brightness gradient information)\ to
output space can be made. As in the case of dominance stripes alone,
the limited receptive field size and output neuron neighbourhood size
causes the neurons to co-operate together only a few at a time, so
that each local patch of neurons contains a complete mapping from
$R^{2}$ to the 2-dimensional output layer.

\begin{figure}
\begin{centering}
\includegraphics[width=7cm]{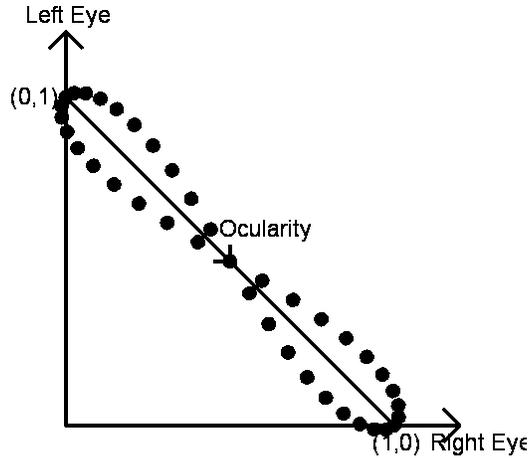}
\par\end{centering}

\caption{Typical neural reference vectors for low resolution input images.}

\label{Fig:Ocular2}
\end{figure}

These results are summarised in figure \ref{Fig:Ocular2} %
 where the pure oscillation back and forth along the ocularity dimension
that occurred in figure \ref{Fig:Ocular1} develops to reveal some
additional degrees of freedom, only one of which is represented in
figure \ref{Fig:Ocular2} (it is perpendicular to the ocularity axis).

\begin{figure}
\begin{centering}
\includegraphics[width=7cm]{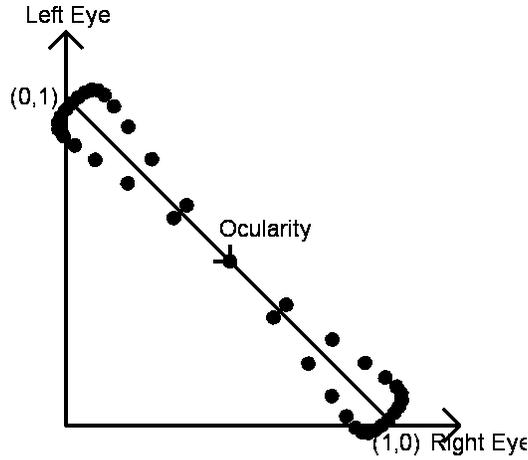}
\par\end{centering}

\caption{Typical neural reference vectors for low resolution input images,
where reduced leakage causes the ocularity to switch abruptly back
and forth.}

\label{Fig:Ocular3}
\end{figure}

If the leakage is reduced then the oscillation back and forth along
the dominance axis tends to be more like a square wave than a sine
wave, in which case figure \ref{Fig:Ocular2} becomes as shown in
figure \ref{Fig:Ocular3} %
 where the neural reference vectors are bunched near to the points
$\left(0,1\right)$ and $\left(1,0\right)$, and explore the additional
degree(s) of freedom at each end of the ocularity axis. In the extreme
case, where the ocularity switches back and forth as a square wave,
the neurons separate into two clusters, one of which responds only
to the left retina's image and the other to the right retina's image.
Furthermore, within each of these clusters, the neurons explore the
additonal degree(s) of freedom that occur within the corresponding
retina's image. Note only one such degree of freedom is represented
in figure \ref{Fig:Ocular3}; it is perpendicular to the ocularity
axis.

The above arguments can be generalised to the case of input images
with fine spatial structure (i.e. lots of high order terms in the
Taylor expansion are required). However, more and more neurons (per
receptive field) are required in order to build a faithful mapping
from input space to a 2-dimensional representation in output space.
For a given number of neurons (per receptive field) a saturation point
will quickly be reached, where the least important detail (from the
point of view of the objective function) is discarded, keeping only
those properties of the input images that best preserve the ability
of the neural network to reconstruct its own input with minimum Euclidean
error (on average).

\section{Simulations\label{Sect:Simulations}}

Two types of training data will be used: synthetic, and natural. Synthetic
data is used in order to demonstrate simple properties of the neural
network, without introducing extraneous detail to complicate the interpretation
of the results. Natural data is used to remove any doubt that the
neural network is capable of producing interesting and useful results
when it encounters data that is more representative of what it might
encounter in the real world.

In section \ref{Sect:Dominance1D} dominance stripes are produced
from a 1-dimensional retina, and in section \ref{Sect:Dominance2D}
these results are generalised to a 2-dimensional retina. In both cases
both synthetic and natural image results are shown. In section \ref{Sect:Orientation}
orientation maps are produced for the case of two retinae trained
with natural images.

\subsection{Dominance Stripes: The 1-Dimensional Case\label{Sect:Dominance1D} }

The purpose of the simulations that are presented in this section
is to demonstrate the emergence of ocular dominance stripes in the
simplest possible realistic case. The results will correspond to the
situation outlined in figure \ref{Fig:Ocular1}.

\subsubsection{Synthetic Training Data}

The purpose of this simulation is to demonstrate the emergence of
ocular dominance stripes, of the type that were shown in figure \ref{Fig:Ocular1},
by presenting a model of the type shown in figure \ref{Fig:Receptive}
with very low-resolution input images. In fact, the resolution is
so low that each image is entirely featureless, so that all the neurons
in a retina have the same input brightness, but the two retinae have
independent input brightnesses. These input images are normalised
by processing them so that they look like the PMD posterior probability
computed by the output layer of another such network; the neighbourhood
size used for this normalisation process was chosen to be the same
as the network's own output layer neighbourhood size.

In the first simulation the parameters used were: network size = $30$,
receptive field size = $9$, output layer neighbourhood size = $5$
(centred on the source neuron), leakage neighbourhood size = $5$
(centred on the source neuron), number of training updates = $2000$,
update step size = $0.01$. For each neuron the leakage probability
had a Gaussian profile centred on the neuron, and the standard deviation
was chosen as $1$, to make the profile fall from $1$ on the source
neuron to $\exp\left(-1/2\right)$ on each of its two closest neighbours.

The update scheme used was a crude gradient following algorithm parameterised
by three numbers which controlled the rate at which the weight vectors,
biasses and reference vectors were updated. These three numbers were
continuously adjusted to ensure that the maximum rate of change (as
measured over all the neurons in the network) of the length of each
weight vector, and also the maximum rate of change of the absolute
value of each bias, was always equal to the requested update step
size; this prescription will adjust the parameter values until they
jitter around in the neighbourhood of their optimum values. The optimum
reference vectors could in principle be completely determined using
equation \ref{Eq:Reference} for each choice of weights and biasses,
but it is not necessary for the reference vectors to keep in precise
synchrony with the weights and biasses. Rather, the reference vectors
were controlled in a similar way to the weight vectors, except that
they used three times the update step size, which made them more agile
than the weights and biasses they were trying to follow.

\begin{figure}
\begin{centering}
\includegraphics[width=7cm]{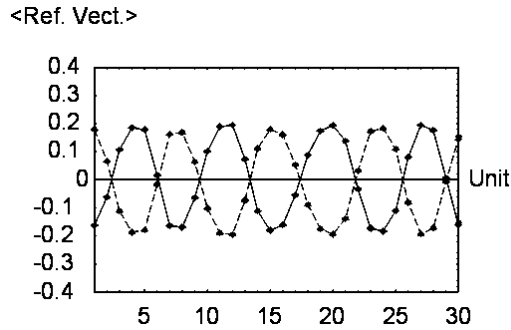}
\par\end{centering}

\caption{1-dimensional dominance stripes after training on synthetic data.}

\label{Fig:Sim1D}
\end{figure}

The ocular dominance stripes that emerge from this simulation are
shown in figure \ref{Fig:Sim1D}. The ocularity for a given neuron
was estimated by computing the average of the absolute deviations
(as measured with respect to the overall mean reference vector component
value, which is zero for the zero mean training data that is used
here) of its reference vector components within its receptive field,
both for the left retina and the right retina. This allows two plots
to be drawn: average value of absolute deviations from the mean in
left retina's receptive field as a function of position across the
network, and similarly the right retina's receptive field. %
 As can be seen in figure \ref{Fig:Sim1D}, these two curves are approximately
periodic, and are in antiphase with each other; this corresponds to
the situation shown in figure \ref{Fig:Ocular1}. The amplitude of
the ocularity curves is less than the $0.5$ that would be required
for the end points of the ocularity dimension to be reached, because
one of the effects of leakage is to introduce a type of elastic tension
between the reference vectors that causes them to contract towards
zero ocularity. Note how the ocular dominance curves have a period
of approximately $7$, which is slightly greater than the output layer
neighbourhood size (which is $5$). In the limit of zero leakage and
infinite receptive field size the period would be equal to the output
layer neighbourhood size, in order to guarantee that a complete set
of processing machinery is contained within each output layer neighbourhood
size; this effect is called completeness.

\begin{figure}
\begin{centering}
\includegraphics[width=7cm]{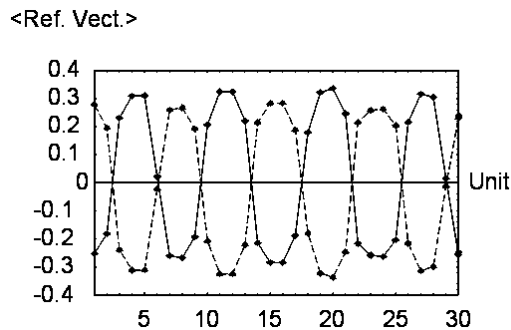}
\par\end{centering}

\caption{1-dimensional square wave dominance stripes after further training
with reduced probability leakage on synthetic data.}

\label{Fig:Sim1D2}
\end{figure}

If the above simulation is continued for a further $2000$ updates
with a reduced leakage, by reducing the standard deviation of the
Gaussian leakage profile from $1$ to $0.5$, then the ocular dominance
curves become more like square waves than sine waves, as shown in
figure \ref{Fig:Sim1D2}; this is similar to the type of situation
that was shown in figure \ref{Fig:Ocular3}, except that the input
images are featureless in this case. %

\subsubsection{Natural Training Data}

\begin{figure}
\begin{centering}
\includegraphics[width=7cm]{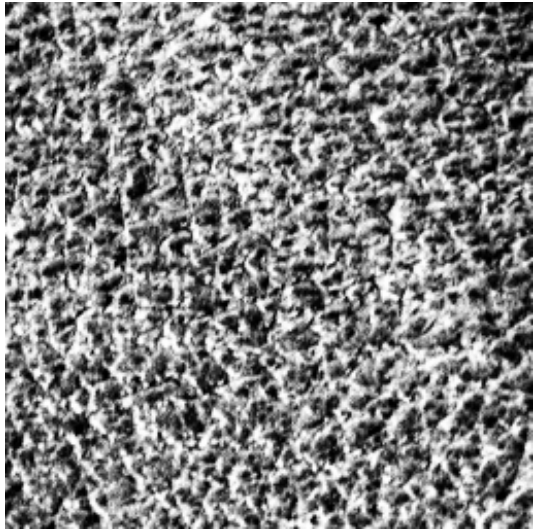}
\par\end{centering}

\caption{Brodatz texture image used as a natural training image.}

\label{Fig:Brodatz}
\end{figure}

\begin{figure}
\begin{centering}
\includegraphics[width=7cm]{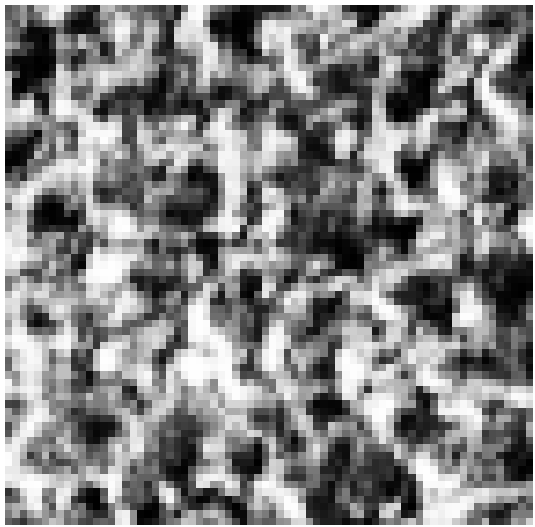}
\par\end{centering}

\caption{Magnified portion of the Brodatz texture training image.}

\label{Fig:BrodatzZoom}
\end{figure}

Figure \ref{Fig:Brodatz} shows the Brodatz texture image \cite{Brodatz1966}
that was used to generate a more realistic training set than was used
in the synthetic simulations described above. %
 Figure \ref{Fig:BrodatzZoom} shows an enlarged portion of figure
\ref{Fig:Brodatz}, where it is clear that the characteristic length
scale of the texture structure is in the range $5-10$ pixels. %
 This is large enough compared to the receptive field size ($9$)
and the output layer neighbourhood size ($5$) that a simulation using
1-dimensional training vectors extracted from this 2-dimensional Brodatz
image will effectively see very low resolution training data, and
should repond approximately as described in figure \ref{Fig:Ocular1}.

\begin{figure}
\begin{centering}
\includegraphics[width=7cm]{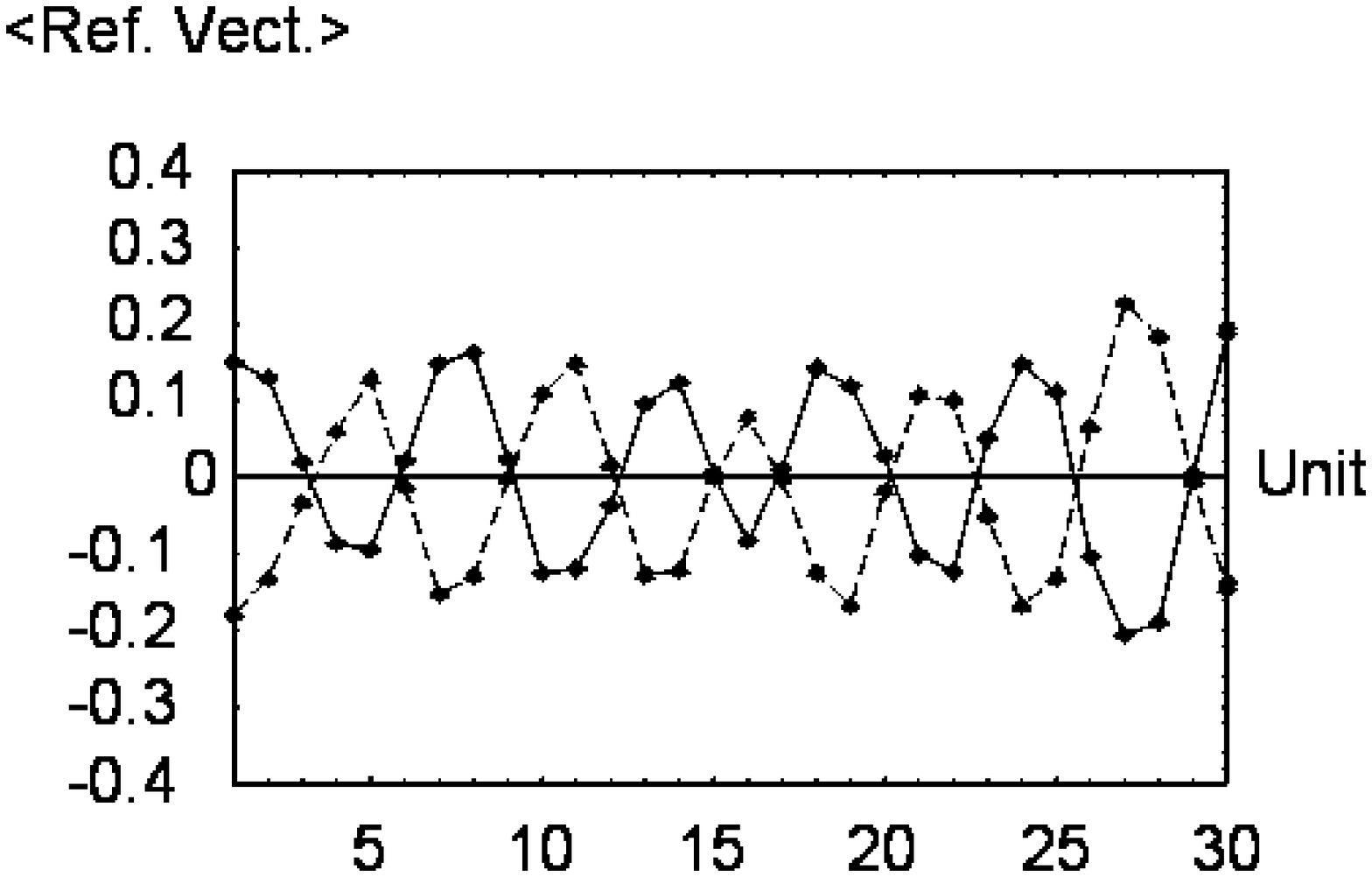}
\par\end{centering}

\caption{1-dimensional dominance stripes after training on natural data.}

\label{Fig:SimReal1D}
\end{figure}

\begin{figure}
\begin{centering}
\includegraphics[width=7cm]{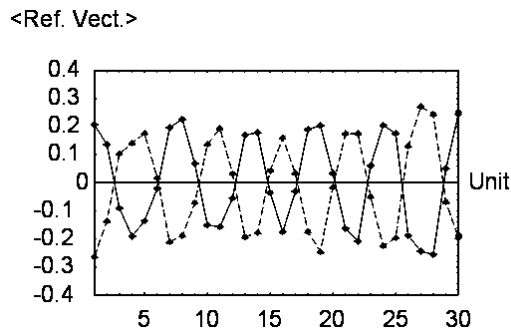}
\par\end{centering}

\caption{1-dimensional square wave dominance stripes after further training
with reduced probability leakage on natural data.}

\label{Fig:SimReal1D2}
\end{figure}

The results corresponding to figure \ref{Fig:Sim1D} and figure \ref{Fig:Sim1D2}
are shown in figure \ref{Fig:SimReal1D} and figure \ref{Fig:SimReal1D2},
respectively. %
 The general behaviour is much the same in the synthetic and Brodatz
cases, except that the depth of the ocularity fluctuations is somewhat
less in the real case, because in the Brodatz case the training data
is not actually featureless within each receptive field.

\subsection{Dominance Stripes:\ The 2-Dimensional Case\label{Sect:Dominance2D}}

This section extends the results of the previous section to the case
of 2-dimensional neural networks. The training schedule(s) used in
the simulations have not been optimised. Usually the update rate is
chosen conservatively (i.e. smaller than it needs to be) to avoid
possible numerical instabilities, and the number of training updates
is chosen to be larger than it needs to be to ensure that convergence
has occurred. It is highly likely that much more efficient training
schedules could be found.

\subsubsection{Synthetic Training Data}

The results that were presented in figure \ref{Fig:Sim1D} may readily
be extended to the case of a 2-dimensional network. The parameters
used were:\
network size = $100\times100$, receptive field size = $3\times3$
(which is artificially small to allow the simulation to run faster),
output layer neighbourhood size = $5\times5$ (centred on the source
neuron), leakage neighbourhood size = $3\times3$ (centred on the
source neuron), number of training updates = $24000$ (dominance stripes
develop quickly, so far fewer than $24000$ training updates could
be used), update step size = $0.001$. For each neuron the leakage
probability had a Gaussian profile centred on the neuron, and the
standard deviations were chosen as $1\times1$, to make the profile
fall from $1$ on the source neuron to $\exp\left(-1/2\right)$ on
each of its four closest neighbours.

\begin{figure}
\begin{centering}
\includegraphics[width=7cm]{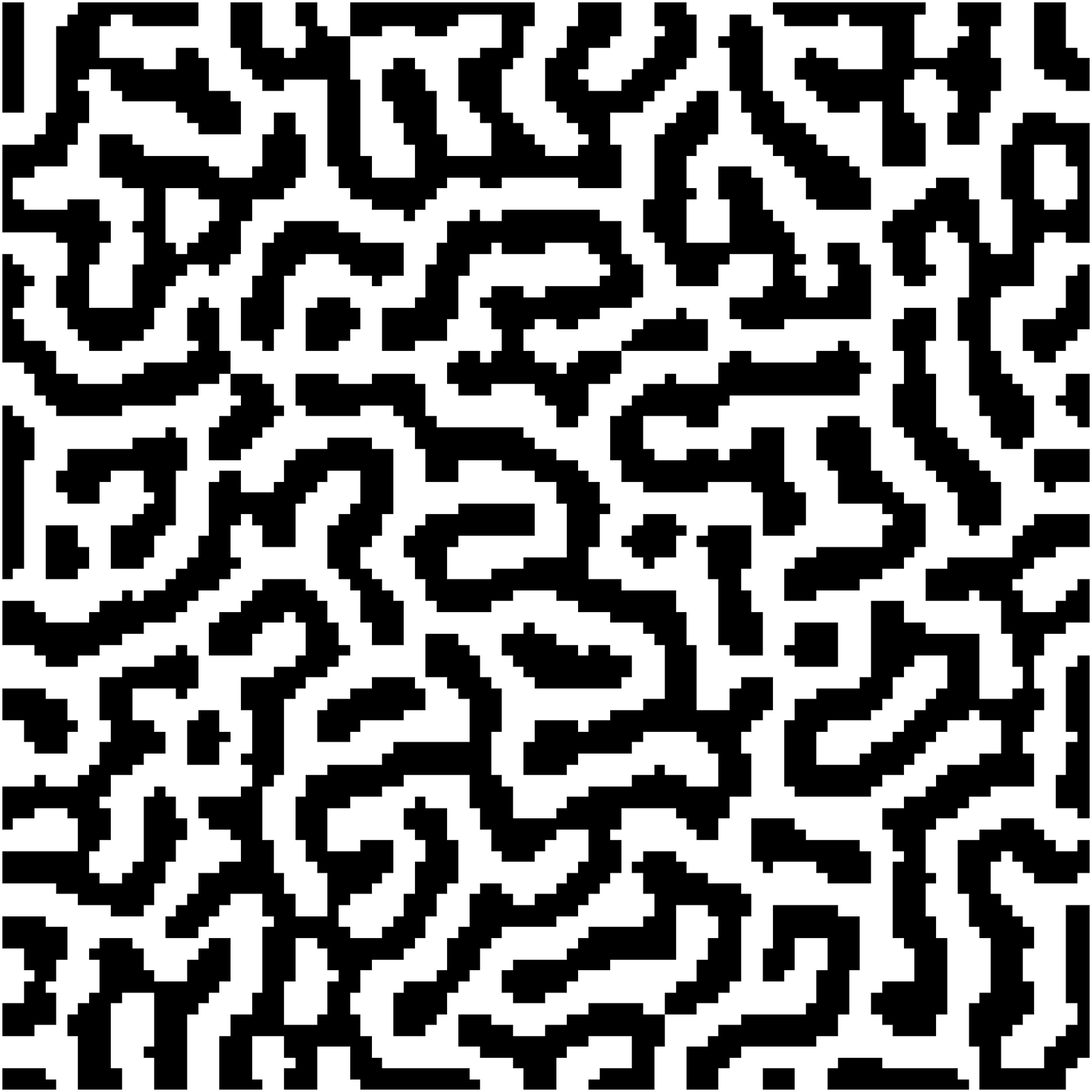}
\par\end{centering}

\caption{2-dimensional dominance stripes after training on synthetic data.}

\label{Fig:Sim2D}
\end{figure}

Apart from the different parameter values, the simulation was conducted
in precisely the same way as in the 1-dimensional case, and the results
for ocular dominance are shown in figure \ref{Fig:Sim2D}, where ocularity
has been quantised as a binary-valued quantity. %
 These results show the characteristic striped structure that is familiar
from experiments on the mammalian visual cortex. The behaviour near
to the boundary depends critically on the interplay between the receptive
field size(s) and the output layer neighbourhood size(s).

\subsubsection{Natural Training Data}

\begin{figure}
\begin{centering}
\includegraphics[width=7cm]{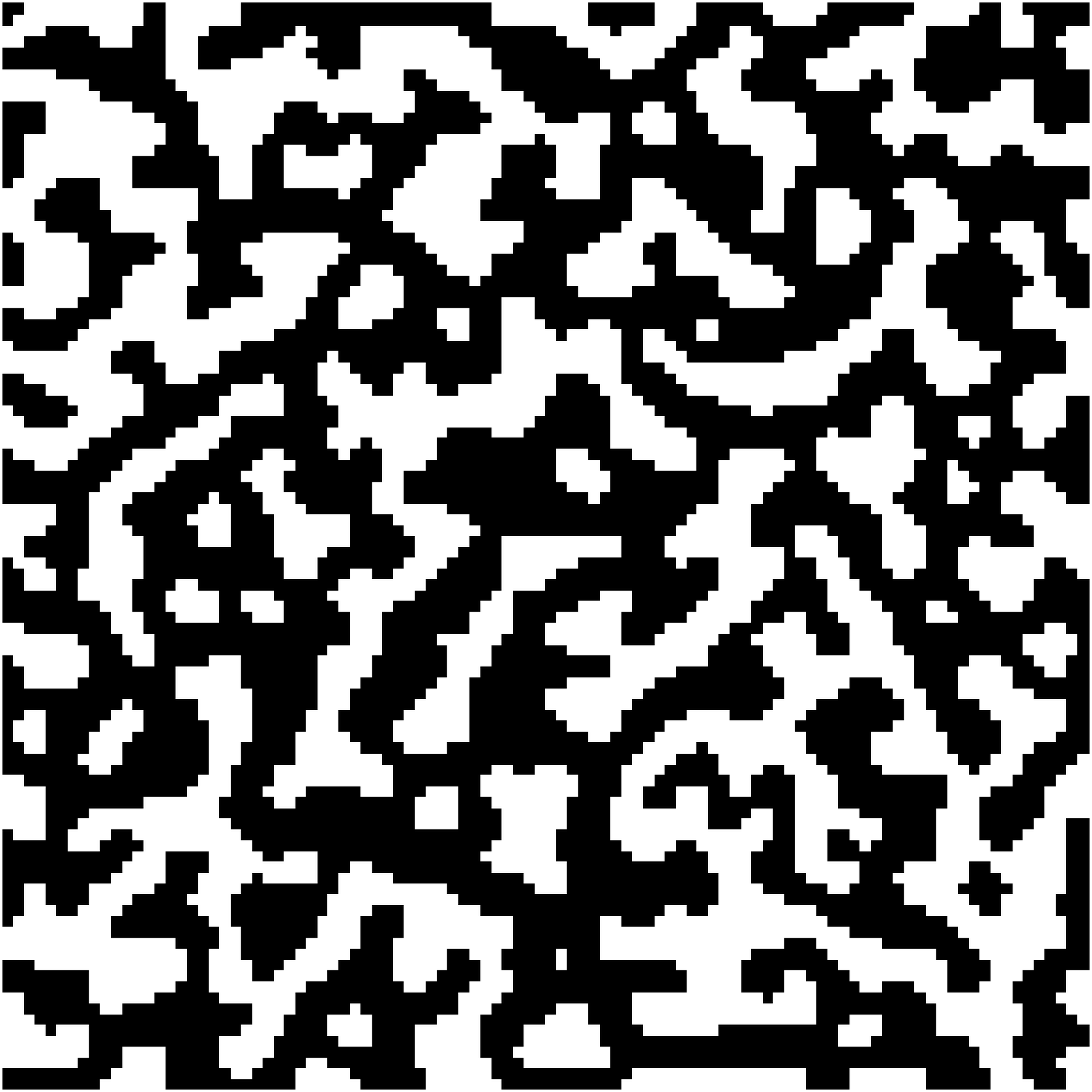}
\par\end{centering}

\caption{2-dimensional dominance stripes after training on natural data.}

\label{Fig:SimReal2D}
\end{figure}

The simulation, whose results were shown in figure \ref{Fig:Sim2D},
may be repeated using the Brodatz image training set shown in figure
\ref{Fig:Brodatz}, to yield the results shown in figure \ref{Fig:SimReal2D}.
 These results are not quite as stripe-like as the results in figure
\ref{Fig:Sim2D}, because in the Brodatz case the training data is
not actually featureless within each receptive field.

\subsection{Orientation Maps\label{Sect:Orientation}}

The purpose of the simulations that are presented in this section
is to demonstrate the emergence of orientation maps in the simplest
possible realistic case. In the case of two retinae, the results will
correspond to the situation outlined in figure \ref{Fig:Ocular2}
(or, at least, a higher dimensional version of that figure).

\subsubsection{Orientation Map (One Retina)}

In this simulation the parameters used were: network size = $30\times30$,
receptive field size = $17\times17$, output layer neighbourhood size
= $9\times9$ (centred on the source neuron), leakage neighbourhood
size = $3\times3$ (centred on the source neuron), number of training
updates = $24000$, update step size = $0.01$. For each neuron the
leakage probability had a Gaussian profile centred on the neuron,
and the standard deviations were chosen as $1\times1$, to make the
profile fall from $1$ on the source neuron to $\exp\left(-1/2\right)$
on each of its four closest neighbours.

Note that both the receptive field size and the output layer neighbourhood
size are substantially larger than in the 2-dimensional dominance
stripe simulations, because many more neurons are required in order
to allow orientation maps to develop than to allow dominance stripes
to develop; in fact it would be preferable to use even larger sizes
than were used here. To limit the computer run time this meant that
the overall size of the neural network had to be reduced from $100\times100$
to $30\times30$. The training set was the Brodatz texture image in
figure \ref{Fig:Brodatz}.

\begin{figure}
\begin{centering}
\includegraphics[width=7cm]{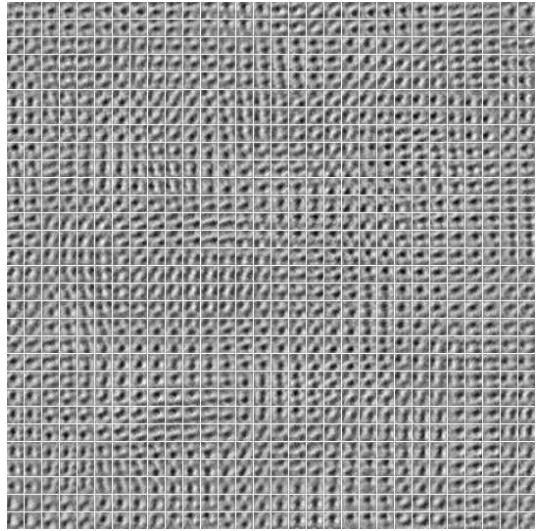}
\par\end{centering}

\caption{Orientation map after training on natural data.}

\label{Fig:Orientation1}
\end{figure}

The results are shown in figure \ref{Fig:Orientation1} %
 where the receptive fields have been gathered together in a montage.
There is a clear swirl-like pattern that is characteristic of orientation
maps. Each local clockwise or anticlockwise swirl typically circulates
around an unoriented region.

\subsubsection{Using the Orientation Map}

\begin{figure}
\begin{centering}
\includegraphics[width=7cm]{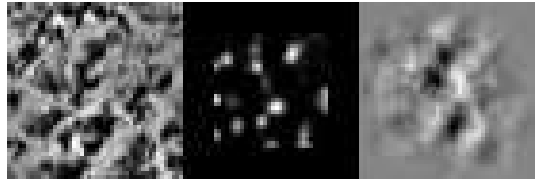}
\par\end{centering}

\caption{Typical input, output and reconstruction produced by the orientation
map.}

\label{Fig:Sparse}
\end{figure}

In figure \ref{Fig:Sparse} the orientation map network shown in figure
\ref{Fig:Orientation1} is used to encode and decode a typical input
image. On the left of figure \ref{Fig:Sparse} the input image (i.e.
$\mathbf{x}$) is shown, in the centre of figure \ref{Fig:Sparse}
the corresponding output (i.e. its PMD posterior probability $\Pr\left(y|\mathbf{x}\right)$)
produced by the orientation map is shown, and on the right of figure
\ref{Fig:Sparse} the corresponding reconstruction (i.e. $\sum_{y=1}^{M}\Pr\left(y|\mathbf{x}\right)\mathbf{x}^{\prime}\left(y\right)$)\ is
shown.

The output consists of a number of isolated {}``activity bubbles''
of posterior probability, and the reconstruction is a low resolution
version of the original input. The form of output is familiar as a
type of {}``sparse coding'' of the input, where only a small fraction
of the neurons participate in encoding a given input (this type of
transformation of the input\ is central to the work that was reported
in \cite{Webber1994}). This type of encoding is very convenient because
it has effectively transformed the input into a small number of constituents
each of which corresponds to an activity bubble, rather than transforming
the input into a representation where the output activity is spread
over all of the neurons, which is thus not easily interpretable as
arising from a small number of constituents.

The reconstruction has a lower resolution than the input because there
are insufficient neurons to faithfully record all the information
that is required to reconstruct the input exactly (e.g. probability
leakage causes neighbouring neurons to have a correlated response,
thus reducing the effective number of neurons that are available).
The featureless region around the edge of the reconstruction is an
artefact, which occurs because fewer neurons (per unit area) contribute
to the reconstruction near the edge of the input array.

\subsubsection{Orientation Map (Two Retinae)}

The above orientation map results may be generalised to the case of
two retinae. The parameter values used were the same, apart from the
standard deviation of the leakage Gaussian which was reduced to $0.5\times0.5$
in order to allow more detailed structure to develop in the adaptive
parameter values of the output neurons. This is necessary because
the presence of two retinae causes dominance stripes to develop, which
allows only half of the neurons to be allocated to each retina, so
a complete repertoire of computing machinery must be forced into half
the number of neurons that were used in the case of one retina.

\begin{figure}
\begin{centering}
\includegraphics[width=7cm]{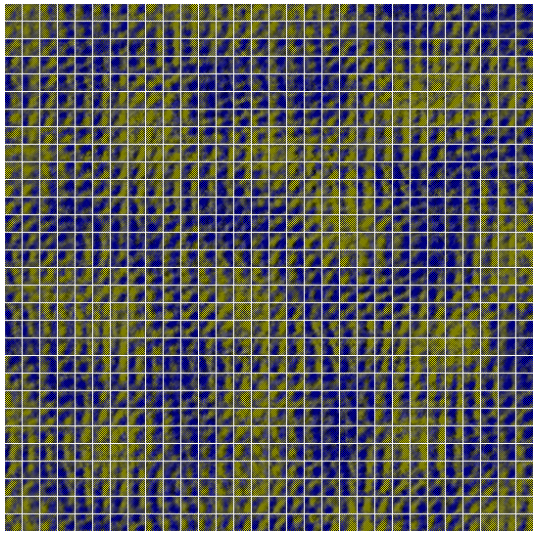}
\par\end{centering}

\caption{Orientation map and dominance stripes after training on natural data.}

\label{Fig:Orientation2}
\end{figure}

The results are shown in figure \ref{Fig:Orientation2} %
 where the receptive fields for the left and right retinae have been
used to create a colour separation in which one retina is coded as
blue and the other as yellow. Within each retina there is a long-scale
periodic fluctuation in overall brightness which corresponds to the
dominance stripes. Within each dominance stripe there is the characteristic
swirl-like pattern of the orientation map. Note that the unoriented
regions typically occur at the centre of dominance stripes, as observed
in the visual cortex; this can be understood intuitively by referring
to figure \ref{Fig:Ocular3}.

A larger simulation would be required in order to accurately estimate
the detailed orientation map as a vector flow field. Such simulations
could be used to verify whether the iso-orientation contours typically
lie perpendicular to the dominance stripe boundaries, as observed
in the visual cortex. The dominance stripe structure that appears
in this simulation is not as distinct as the stripes in figure \ref{Fig:SimReal2D}.
This is not a fundamental problem, but rather it is a result of the
limited size of computer simulation that could be run in a reasonable
length of time. It should also be noted that the dominance stripes
that are observed in the visual cortex are sometimes more blob-like
than stripe-like \cite{Swindale1996}, so it is pleasing that different
choices of parameter value should yield a variety of degrees of stripiness
in our simulations.

\section{Conclusions\label{Sect:Conclusions}}

This paper has shown how folded Markov chains (FMCs) \cite{Luttrell1994a}
can be combined with partitioned mixture distributions (PMDs) \cite{Luttrell1994b}
to yield a class of self-organising neural networks that has many
of the properties that are observed in the mammalian visual cortex
\cite{Goodhill1992,Swindale1996}, which are thus called visual cortex
networks (VICON). These neural networks differ from previous models
of the visual cortex, insofar as they model the neuron behaviour in
terms of their individual firing events, and operate in the real space
of input images rather than a hand-crafted abstract space, and the
use of Bayesian methods makes the nature of the network's computations
clearer than in the case where the network behaviour is simply postulated.
When the neural network structure (e.g. receptive field size) parameters
are appropriately chosen, dominance stripes and orientation maps emerge
naturally when the network is trained on a natural image (e.g. a Brodatz
texture image).

These results show how this type of network is capable of self-organising
its internal parameters in familiar ways when trained on data from
multiple sources (actually, only two sources in the case of the visual
cortex-like network). The same network objective function could be
used when an arbitrary number of data sources is presented, and it
is anticipated that it would lead to analogous results.

An extension of the network objective function to the case where sets
of multiple neural firing events are considered has been published
\cite{Luttrell1995a,Luttrell1995b}, and an extension to the case
of a multilayer network has been published \cite{Luttrell1996}. When
combined, these extensions could be applied to the problem of the
processing of data from multiple sensors (i.e. data fusion).

\appendix

\section{Bayesian PMD\label{Sect:BayesPMD}}

In this section a fully Bayesian interpretation of a partitioned mixture
distribution (PMD) will be presented.

Consider the general problem of computing a posterior probability
$\Pr\left(y|\mathbf{x}\right)$ over classes $y$ given an input vector
$\mathbf{x}$. If there is more than one model $k$ then $\Pr\left(y|\mathbf{x}\right)$
is given by a marginal PDF \begin{equation}
\Pr\left(y|\mathbf{x}\right)=\sum_{k}\Pr\left(y,k|\mathbf{x}\right)\end{equation}
 where $\Pr\left(y,k|\mathbf{x}\right)$ is the joint PDF of class
$y$ and model $k$ given an input vector $\mathbf{x}$. Bayes' theorem
may be used to rewrite this as follows \begin{eqnarray}
\Pr\left(y,k|\mathbf{x}\right) & = & \frac{\Pr\left(y,k,\mathbf{x}\right)}{\Pr\left(\mathbf{x}\right)}\nonumber \\
 & = & \frac{\Pr\left(y|k,\mathbf{x}\right)\Pr\left(k,\mathbf{x}\right)}{\Pr\left(\mathbf{x}\right)}\nonumber \\
 & = & \Pr\left(y|k,\mathbf{x}\right)\Pr\left(k\right)\end{eqnarray}
 where $\Pr\left(k,\mathbf{x}\right)=\Pr\left(k\right)\Pr\left(\mathbf{x}\right)$
(i.e. independence of model $k$ and data vector $\mathbf{x}$)\ has
been assumed in the last step. Thus the posterior probability $\Pr\left(y|\mathbf{x}\right)$
may be written as \begin{equation}
\Pr\left(y|\mathbf{x}\right)=\sum_{k}\Pr\left(y|k,\mathbf{x}\right)\Pr\left(k\right)\label{Eq:BayesMix}\end{equation}
 Assume that there are $M$ models, and that the prior probabilities
$\Pr\left(k\right)$ of the various models are equal, so that $\Pr\left(k\right)=\frac{1}{M}$,
in which case the posterior probability reduces to \begin{equation}
\Pr\left(y|\mathbf{x}\right)=\frac{1}{M}\sum_{k=1}^{M}\Pr\left(y|k,\mathbf{x}\right)\end{equation}
 which is an average of $M$ contributing posterior probabilities
(one from each of the contributing models). The PMD posterior probability
in equation \ref{Eq:PosteriorPMD} is a special case of this result.

More generally, the prior probabilities $\Pr\left(k\right)$ are $k$-dependent,
and might be chosen in some optimal fashion to best handle the training
set. The simplest way of determining an optimal $\Pr\left(k\right)$
is to minimise $D$ with respect to $\Pr\left(k\right)$; this merely
extends the space in which $D$ is optimised to include more of the
parameters inside $\Pr\left(y|\mathbf{x}\right)$.

\section{Optimal Solutions\label{Sect:Optimal}}

In this section the the objective function $D$\ will be minimised
in the case where the input space consists of one or more subspaces,
within each of which all of the input vector components have the same
value. In the language of imaging sensors, these special cases correspond
to each sensor viewing a featureless scene (i.e. all pixels having
the same brightness value), which is effectively the lowest order
term in a Taylor expansion of the spatial variation of pixel brightness
values. This might not appear to be an interesting scenario to consider,
but it leads to a highly non-trivial optimal network behaviour when
$D$ is minimised. More complicated input statistics leads to even
more complicated optimal network behaviour, so only the simplest case
described above will be considered at first.

\subsection{One Input Subspace}

This may be used to optimise the network for a single sensor viewing
a featureless scene. For a $d$-dimensional input space $\Pr\left(\mathbf{x}\right)$
is thus given by \begin{equation}
\Pr\left(\mathbf{x}\right)=\Pr\left(x_{1}\right)\prod_{i=2}^{d}\delta\left(x_{i}-x_{1}\right)\end{equation}
 whence the objective function $D$ in equation \ref{Eq:ObjectiveSimplified}
reduces to

\begin{equation}
D=2d\int dx_{1}\Pr\left(x_{1}\right)\sum_{y=1}^{M}\Pr\left(y|x_{1}\right)\left(x_{1}-x_{1}^{\prime}\left(y\right)\right)^{2}\end{equation}
 This is $d$ times the objective function for a 1-dimensional soft
scalar quantiser which encodes inputs in $x_{1}$-space whose PDF
is $\Pr\left(x_{1}\right)$.

\subsection{Two Input Subspaces}

This may be used to optimise the network for a pair of sensors each
of which views a featureless scene, and which are possibly correlated
with each other. The one input subspace case above can readily be
generalised to more input subspaces. Let the $d$-dimensional input
space be split into two $\frac{d}{2}$-dimensional subspaces, where
$\Pr\left(\mathbf{x}\right)$ is given by \begin{equation}
\Pr\left(\mathbf{x}\right)=\Pr\left(x_{1},x_{2}\right)\prod_{i=2}^{\frac{d}{2}}\delta\left(x_{2\bar{\imath}-1}-x_{1}\right)\delta\left(x_{2\bar{\imath}}-x_{2}\right)\end{equation}
 where one of the subspaces consists of the odd-numbered components,
and the other the even-numbered components of the input vector (this
particular ordering of the components is not important). Whence the
objective function $D$ in equation \ref{Eq:ObjectiveSimplified}
reduces to \begin{eqnarray}
D & = & d\int dx_{1}dx_{2}\Pr\left(x_{1},x_{2}\right)\sum_{y=1}^{M}\Pr\left(y|x_{1},x_{2}\right)\nonumber \\
 &  & \times\left(\left(x_{1}-x_{1}^{\prime}\left(y\right)\right)^{2}+\left(x_{2}-x_{2}^{\prime}\left(y\right)\right)^{2}\right)\end{eqnarray}
 This is $\frac{d}{2}$ times the objective function for a 2-dimensional
soft vector quantiser which encodes inputs in $\left(x_{1},x_{2}\right)$-space
whose PDF is $\Pr\left(x_{1},x_{2}\right)$. This result generalises
in the obvious way to a larger number of input subspaces.

\subsection{PMD Posterior Probability}

In the above special cases each neuron potentially responds to all
of the components of the input vector. If this were to be built in
hardware, then each neuron would have a number of inputs equal to
the dimensionality of the input space, which becomes unwieldy if the
input space had a high dimensionality (e.g. an image). For high-dimensional
inputs it is sensible to limit the number of inputs to each neuron,
which can readily be implemented by imposing a finite-sized receptive
field on the input of each neuron, such that it can respond only to
a limited subset of all of the input vector components. This constraint
will prevent the ideal vector quantiser solutions from being obtained,
so the purpose of this section is to derive the constrained optimal
solution. Note that this type of input is a special case of the type
of solution that would be obtained by adding a {}``wire-length''
penalty term to the objective function in order to penalise the connection
of a neuron to too many input components.

Even if receptive fields are used to restrict the length of the input
connections, the posterior probability $\Pr\left(y|\mathbf{x}\right)$
effectively needs long-range lateral connections between the output
neurons in order to implement the normalisation condition $\sum_{y=1}^{M}\Pr\left(y|\mathbf{x}\right)=1$.
The simplest example of this is the standard vector quantiser, whose
winner-take-all property requires that all neurons are laterally connected
to all other neurons \textit{even if} each of them has only a finite-sized
receptive field. A partitioned mixture distribution (PMD) posterior
probability, in which the posterior probability $\Pr\left(y|\mathbf{x}\right)$
is only locally connected, can be used to ensure that all the connections
in the network are local (see section \ref{Sect:PostProbPMD}).

\subsubsection{Receptive Fields}

Write the input vector as $\mathbf{x=}\left(\mathbf{\tilde{x}}\left(y\right),\mathbf{\bar{x}}\left(y\right)\right)$
where $\mathbf{\tilde{x}}\left(y\right)$ is the part of $\mathbf{x}$
that lies within the receptive field of neuron $y$, and, for simplicity,
assume that the receptive field used for $\mathbf{x}^{\prime}\left(y\right)$
is chosen to be the same as that for $\mathbf{\tilde{x}}\left(y\right)$,
and that all receptive fields see the same number $w$ of input components.
Because the input vector is split into two$\,$subspaces as $\mathbf{x}=\left(\mathbf{x}_{1},\mathbf{x}_{2}\right)$,
its decompositon as $\left(\mathbf{\tilde{x}}\left(y\right),\mathbf{\bar{x}}\left(y\right)\right)$
may similarly be split into two$\,$subspaces as $\mathbf{\tilde{x}}\left(y\right)=\left(\mathbf{\tilde{x}}_{1}\left(y\right),\mathbf{\tilde{x}}_{2}\left(y\right)\right)$
and $\mathbf{\bar{x}}\left(y\right)=\left(\mathbf{\bar{x}}_{1}\left(y\right),\mathbf{\bar{x}}_{2}\left(y\right)\right)$.
Use the orthogonality of $\mathbf{\tilde{x}}\left(y\right)$ and $\mathbf{\bar{x}}\left(y\right)$
to write (for $i=1,2$) \[
\left\Vert \mathbf{\bar{x}}_{i}\left(y\right)+\mathbf{\tilde{x}}_{i}\left(y\right)-\mathbf{x}_{i}^{\prime}\left(y\right)\right\Vert ^{2}=\left\Vert \mathbf{\bar{x}}_{i}\left(y\right)\right\Vert ^{2}+\left\Vert \mathbf{\tilde{x}}_{i}\left(y\right)-\mathbf{x}_{i}^{\prime}\left(y\right)\right\Vert ^{2}\]
 and simplify $D$ in equation \ref{Eq:ObjectiveSimplified} thus

\begin{eqnarray}
D & = & 2\int d\mathbf{x}_{1}d\mathbf{x}_{2}\Pr\left(\mathbf{x}_{1},\mathbf{x}_{2}\right)\sum_{y}\Pr\left(y|\mathbf{x}_{1},\mathbf{x}_{2}\right)\nonumber \\
 &  & \times\left(\begin{array}{c}
\left\Vert \mathbf{\bar{x}}_{1}\left(y\right)\right\Vert ^{2}+\left\Vert \mathbf{\bar{x}}_{2}\left(y\right)\right\Vert ^{2}\\
+\left\Vert \mathbf{\tilde{x}}_{1}\left(y\right)-\mathbf{x}_{1}^{\prime}\left(y\right)\right\Vert ^{2}+\left\Vert \mathbf{\tilde{x}}_{2}\left(y\right)-\mathbf{x}_{2}^{\prime}\left(y\right)\right\Vert ^{2}\end{array}\right)\end{eqnarray}
 There are two terms to consider.
\begin{enumerate}
\item $\left\Vert \mathbf{\bar{x}}_{1}\left(y\right)\right\Vert ^{2}+\left\Vert \mathbf{\bar{x}}_{2}\left(y\right)\right\Vert ^{2}$.
This is the contribution from \textit{outside} the $y^{th}$ receptive
field, which is the $L_{2}$ norm of those components of the input
vector that lie outside the $y^{th}$ receptive field.
\item $\left\Vert \mathbf{\tilde{x}}_{1}\left(y\right)-\mathbf{x}_{1}^{\prime}\left(y\right)\right\Vert ^{2}+\left\Vert \mathbf{\tilde{x}}_{2}\left(y\right)-\mathbf{x}_{2}^{\prime}\left(y\right)\right\Vert ^{2}$:
This is the contribution from \textit{inside} the $y^{th}$ receptive
field, which is the $L_{2}$ norm of those components of the error
vector (i.e. input minus reconstruction)\
that lie inside the $y^{th}$ receptive field.
\end{enumerate}

\subsubsection{Simplify the $\left\Vert \mathbf{\bar{x}}_{1}\left(y\right)\right\Vert ^{2}+\left\Vert \mathbf{\bar{x}}_{2}\left(y\right)\right\Vert ^{2}$
Term}

$\left\Vert \mathbf{\bar{x}}_{1}\left(y\right)\right\Vert ^{2}+\left\Vert \mathbf{\bar{x}}_{2}\left(y\right)\right\Vert ^{2}$
is the $L_{2}$ norm of those components of the input vector that
lie outside the $y^{th}$ receptive field, which is known once the
input vector is specified. Furthermore, because of the assumed input
PDF (i.e. all input components in each subspace have the same value),
together with the assumed receptive field prescription (i.e. all receptive
fields are the same size $w$), this $L_{2}$ norm is independent
of $y$ given that $\mathbf{x}$ is known, so this term has the following
contribution to $D$\begin{equation}
D=\left(d-w\right)\left(\int dx_{1}\Pr\left(x_{1}\right)x_{1}^{2}+\int dx_{2}\Pr\left(x_{2}\right)x_{2}^{2}\right)\label{Eq:Doutside}\end{equation}
 where $d-w$ is the number of input components that lie \textit{outside}
each receptive field.

\subsubsection{Simplify the $\left\Vert \mathbf{\tilde{x}}_{1}\left(y\right)-\mathbf{x}_{1}^{\prime}\left(y\right)\right\Vert ^{2}+\left\Vert \mathbf{\tilde{x}}_{2}\left(y\right)-\mathbf{x}_{2}^{\prime}\left(y\right)\right\Vert ^{2}$
Term}

Assume that $\Pr\left(y|\mathbf{x}_{1},\mathbf{x}_{2}\right)$ has
the PMD form of a sum over mixture distribution posterior probabilities
(as described in section \ref{Sect:PostProbPMD}), so that \begin{eqnarray}
\Pr\left(y|\mathbf{x}_{1},\mathbf{x}_{2}\right) & = & \frac{1}{M}\sum_{y^{\prime}\in\mathcal{N}^{-1}\left(y\right)}\Pr\left(y|\mathbf{x}_{1},\mathbf{x}_{2}\mathbf{;}y^{\prime}\right)\nonumber \\
 & = & \frac{1}{M}\, Q\left(\mathbf{\tilde{x}}_{1},\mathbf{\tilde{x}}_{2}\mathbf{|}y\right)\sum_{y^{\prime}\in\mathcal{N}^{-1}\left(y\right)}\frac{1}{\sum_{y^{\prime\prime}\in\mathcal{N}\left(y^{\prime}\right)}Q\left(\mathbf{\tilde{x}}_{1},\mathbf{\tilde{x}}_{2}\mathbf{|}y^{\prime\prime}\right)}\end{eqnarray}
 The overall receptive field that effects the value of $\Pr\left(y|\mathbf{x}_{1},\mathbf{x}_{2}\right)$
(for a given $y$) may be read off this expression. Thus $\mathbf{\tilde{x}}\left(y^{\prime\prime}\right)$
comprises those components of $\mathbf{x}$ that lie within the receptive
field of neuron $y^{\prime\prime}$, and the $\sum_{y^{\prime}\in\mathcal{N}^{-1}\left(y\right)}\frac{1}{\sum_{y^{\prime\prime}\in\mathcal{N}\left(y^{\prime}\right)}\left(\cdots\right)}$
operation compounds these $\mathbf{\tilde{x}}\left(y^{\prime\prime}\right)$
so that the overall set of components of $\mathbf{x}$ that are needed
for the purposes of calculating $\Pr\left(y|\mathbf{x}_{1},\mathbf{x}_{2}\right)$
is given by (using a somewhat cavalier notation) \begin{equation}
\mathbf{\tilde{X}}\left(y\right)\equiv\bigcup_{y^{\prime}\in\mathcal{N}^{-1}\left(y\right)}\bigcup_{y^{\prime\prime}\in\mathcal{N}\left(y^{\prime}\right)}\mathbf{\tilde{x}}\left(y^{\prime\prime}\right)\end{equation}
 The individual $\Pr\left(y|\mathbf{x}_{1},\mathbf{x}_{2}\mathbf{;}y^{\prime}\right)$
that contribute to $\Pr\left(y|\mathbf{x}_{1},\mathbf{x}_{2}\right)$
each depend on a smaller set of components of $\mathbf{x}$ than the
full $\Pr\left(y|\mathbf{x}_{1},\mathbf{x}_{2}\right)$, because there
is one less summation over a $y$ variable. However, it is convenient,
and imposes no constraint, to use the full set of components thus
\begin{equation}
\Pr\left(y|\mathbf{x}_{1},\mathbf{x}_{2};y^{\prime}\right)=\Pr\left(y|\mathbf{\tilde{X}}_{1}\left(y\right),\mathbf{\tilde{X}}_{2}\left(y\right);y^{\prime}\right)\end{equation}
 The $y$ and $y^{\prime}$ summations can be interchanged using $\sum_{y=1}^{M}\sum_{y^{\prime}\in\mathcal{N}^{-1}\left(y\right)}\left(\cdots\right)=\sum_{y^{\prime}=1}^{M}\sum_{y\in\mathcal{N}\left(y^{\prime}\right)}\left(\cdots\right)$,
whence the contribution to $D$ is \begin{eqnarray}
D & = & \frac{2}{M}\int d\mathbf{x}_{1}\, d\mathbf{x}_{2}\Pr\left(\mathbf{x}_{1},\mathbf{x}_{2}\right)\sum_{y^{\prime}=1}^{M}\sum_{y\in\mathcal{N}\left(y^{\prime}\right)}\Pr\left(y|\mathbf{\tilde{X}}_{1}\left(y\right),\mathbf{\tilde{X}}_{2}\left(y\right);y^{\prime}\right)\nonumber \\
 &  & \times\left(\left\Vert \mathbf{\tilde{x}}_{1}\left(y\right)-\mathbf{x}_{1}^{\prime}\left(y\right)\right\Vert ^{2}+\left\Vert \mathbf{\tilde{x}}_{2}\left(y\right)-\mathbf{x}_{2}^{\prime}\left(y\right)\right\Vert ^{2}\right)\end{eqnarray}
 Because the components of $\mathbf{\tilde{x}}_{i}\left(y\right)$
are a subset of the components of $\mathbf{\tilde{X}}_{i}\left(y\right)$(for
$i=1,2$), $\Pr\left(\mathbf{x}_{1},\mathbf{x}_{2}\right)$ can be
marginalised to yield \begin{eqnarray}
D & = & \frac{2}{M}\sum_{y^{\prime}=1}^{M}\sum_{y\in\mathcal{N}\left(y^{\prime}\right)}\int d\mathbf{\tilde{X}}_{1}\left(y\right)\, d\mathbf{\tilde{X}}_{2}\left(y\right)\Pr\left(\mathbf{\tilde{X}}_{1}\left(y\right),\mathbf{\tilde{X}}_{2}\left(y\right)\right)\nonumber \\
 &  & \times\Pr\left(y|\mathbf{\tilde{X}}_{1}\left(y\right),\mathbf{\tilde{X}}_{2}\left(y\right);y^{\prime}\right)\nonumber \\
 &  & \times\left(\left\Vert \mathbf{\tilde{x}}_{1}\left(y\right)-\mathbf{x}_{1}^{\prime}\left(y\right)\right\Vert ^{2}+\left\Vert \mathbf{\tilde{x}}_{2}\left(y\right)-\mathbf{x}_{2}^{\prime}\left(y\right)\right\Vert ^{2}\right)\end{eqnarray}
 Because $\Pr\left(\mathbf{x}_{1},\mathbf{x}_{2}\right)$ specifies
that all of the components in each subspace are the same, this contribution
to $D$ may be simplified to \begin{eqnarray}
D & = & \frac{w}{M}\sum_{y^{\prime}=1}^{M}\sum_{y\in\mathcal{N}\left(y^{\prime}\right)}\int dx_{1}\, dx_{2}\Pr\left(x_{1},x_{2}\right)\Pr\left(y|x_{1},x_{2};y^{\prime}\right)\nonumber \\
 &  & \times\left(\left(x_{1}-x_{1}^{\prime}\left(y\right)\right)^{2}+\left(x_{2}-x_{2}^{\prime}\left(y\right)\right)^{2}\right)\label{Eq:Dinside}\end{eqnarray}

\subsubsection{Periodic Optimal Solutions}

Combining the results from outside (equation \ref{Eq:Doutside}) and
inside (equation \ref{Eq:Dinside}) the receptive fields yields finally
\begin{eqnarray}
D & = & \left(d-w\right)\left(\int dx_{1}\Pr\left(x_{1}\right)x_{1}^{2}+\int dx_{2}\Pr\left(x_{2}\right)x_{2}^{2}\right)\nonumber \\
 &  & +\frac{w}{M}\sum_{y^{\prime}=1}^{M}\sum_{y\in\mathcal{N}\left(y^{\prime}\right)}\int dx_{1}\, dx_{2}\Pr\left(x_{1},x_{2}\right)\Pr\left(y|x_{1},x_{2};y^{\prime}\right)\nonumber \\
 &  & \times\left(\left(x_{1}-x_{1}^{\prime}\left(y\right)\right)^{2}+\left(x_{2}-x_{2}^{\prime}\left(y\right)\right)^{2}\right)\end{eqnarray}
 The first of these terms is constant, so it may be ignored insofar
as network optimisation is concerned. The second term is much more
interesting. It is the sum of the objective functions of a large number
of 2-dimensional soft vector quantisers. However, these objective
functions cannot be optimised independently of each other, because
the posterior probabilities $\Pr\left(y|x_{1},x_{2};y^{\prime}\right)$
force the neurons to share parameters with each other.

Drop the constant term, and interchange the order of summation to
obtain

\begin{eqnarray}
D & = & w\sum_{y=1}^{M}\int dx_{1}\, dx_{2}\Pr\left(x_{1},x_{2}\right)\Pr\left(y|x_{1},x_{2}\right)\nonumber \\
 &  & \times\left(\left(x_{1}-x_{1}^{\prime}\left(y\right)\right)^{2}+\left(x_{2}-x_{2}^{\prime}\left(y\right)\right)^{2}\right)\end{eqnarray}
 where $\Pr\left(y|x_{1},x_{2}\right)$ is the PMD posterior probability
given by \begin{equation}
\Pr\left(y|x_{1},x_{2}\right)=\frac{1}{M}\sum_{y^{\prime}\in\mathcal{N}^{-1}\left(y\right)}\Pr\left(y|x_{1},x_{2};y^{\prime}\right)\end{equation}
 Now suppose that $\Pr\left(y|x_{1},x_{2}\right)$ and $x_{i}^{\prime}\left(y\right)$
have the periodicity property \begin{eqnarray}
\Pr\left(y+m|x_{1},x_{2}\right) & = & \Pr\left(y|x_{1},x_{2}\right)\nonumber \\
x_{i}^{\prime}\left(y+m\right) & = & x_{i}^{\prime}\left(y\right)\end{eqnarray}
 where the fact that $y$ is restricted to $1\leq y\leq M$ has been
ignored for simplicity, then $D$ can be simplified thus (again, ignoring
the fact that $y$ is restricted to $1\leq y\leq M$) \begin{eqnarray}
D & = & w\sum_{y_{0}=0}^{\frac{M}{m}-1}\sum_{y=my_{0}+1}^{m\left(y_{0}+1\right)}\int dx_{1}\, dx_{2}\Pr\left(x_{1},x_{2}\right)\Pr\left(y|x_{1},x_{2}\right)\nonumber \\
 &  & \times\left(\left(x_{1}-x_{1}^{\prime}\left(y\right)\right)^{2}+\left(x_{2}-x_{2}^{\prime}\left(y\right)\right)^{2}\right)\nonumber \\
 & = & w\sum_{y=1}^{m}\int dx_{1}\, dx_{2}\Pr\left(x_{1},x_{2}\right)\frac{M}{m}\Pr\left(y|x_{1},x_{2}\right)\nonumber \\
 &  & \times\left(\left(x_{1}-x_{1}^{\prime}\left(y\right)\right)^{2}+\left(x_{2}-x_{2}^{\prime}\left(y\right)\right)^{2}\right)\end{eqnarray}
 where $\frac{M}{m}\sum_{y=1}^{m}\Pr\left(y|x_{1},x_{2}\right)=1$
follows from $\sum_{y=1}^{M}\Pr\left(y|x_{1},x_{2}\right)=1$ and
the periodicity property, so $\frac{M}{m}\Pr\left(y|x_{1},x_{2}\right)$
serves as a posterior probability for $1\leq y\leq m$.

This demonstrates that \textit{if} the optimal solution is periodic,
with period $m$, then the objective function is proportional to the
objective function for a 2-dimensional soft vector quantiser with
$m$ neurons. Note that thus far nothing has been said about the actual
value of $m$; its optimal value depends on the interplay between
the receptive field size(s), the output layer neighbourhood size(s),
and the leakage neighbourhood size(s). Because this type of periodic
solution is essentially a set of overlapping $m$ neuron soft vector
quantisers, each set of $m$ neurons will typically exhibit the properties
of such quantisers. In particular this means that each set of $m$
neurons will have the means to encode and (approximately) reconstruct
those components of the input vector that it sees via its receptive
fields.

This type of solution is the archetype for orientation maps, where
the neurons arrange their properties so that each local patch (corresponding
to the $m$ neurons in the periodic solution derived above) has the
means to encode whatever orientation of object it sees via its receptive
fields. The full derivation of an orientation map would require a
more sophisticated analysis than the simple 1-dimensional case derived
above.

\end{document}